\documentclass[sigconf]{acmart}

\renewcommand\footnotetextcopyrightpermission[1]{} 

\usepackage{xcolor}
\usepackage{epsfig}
\usepackage{graphicx}
\usepackage{amsmath}
\usepackage{enumitem}
\usepackage{sidecap}

\usepackage[font=footnotesize,labelfont=bf]{caption}

\usepackage{multirow}
\usepackage{tabularx}
\usepackage{dsfont}
\usepackage{url}
\usepackage[tight]{subfigure}
\usepackage{color}
\usepackage{subfigure}
\usepackage{amsthm}
\usepackage[export]{adjustbox}
\usepackage{comment}

\usepackage[utf8]{inputenc} 
\usepackage[T1]{fontenc}    
\usepackage{url}            
\usepackage{booktabs}       
\usepackage{amsfonts}       
\usepackage{nicefrac}       
\usepackage{microtype}      
\usepackage{pifont}
\usepackage{textcomp}
\usepackage{tikz}

\definecolor{dg}{rgb}{0,0.694,0.298}
\definecolor{purple}{rgb}{0.4,0.176,0.569}
\definecolor{tabgray}{rgb}{0.85,0.85,0.85}

\usepackage{pifont}
\newcommand{\topone}[1]{\textbf{\textcolor{red}{#1}}}

\newcommand{\figref}[1]{Fig.~\ref{#1}}

\newcommand{\secref}[1]{Sec.~\ref{#1}}
\newcommand{\tableref}[1]{Table~\ref{#1}}

\usepackage{xspace}
\makeatletter
\DeclareRobustCommand\onedot{\futurelet\@let@token\@onedot}
\def\@onedot{\ifx\@let@token.\else.\null\fi\xspace}
\def\eg{\emph{e.g}\onedot} 
\def\ie{\emph{i.e}\onedot}

\makeatother

\AtBeginDocument{%
  \providecommand\BibTeX{{%
    \normalfont B\kern-0.5em{\scshape i\kern-0.25em b}\kern-0.8em\TeX}}}

\setcopyright{acmcopyright}
\copyrightyear{2018}
\acmYear{2018}
\acmDOI{XXXXXXX.XXXXXXX}

\acmPrice{15.00}
\acmISBN{978-1-4503-XXXX-X/18/06}

\acmSubmissionID{2629}

\begin{document}

\title{Learning Restoration is Not Enough: Transfering Identical Mapping for Single-Image Shadow Removal}

\author{Xiaoguang Li$^{1}$, \ Qing Guo$^{2}$, \  Pingping Cai$^{1}$, \ Wei Feng$^{3}$, \ Ivor Tsang$^{2}$, \ Song Wang$^{1}$}

\affiliation{
\institution{$^{1}$ University of South Carolina}
\country{USA}
}

\affiliation{
\institution{$^{2}$ Center for Frontier AI Research (CFAR), A*STAR}
\country{Singapore}
}

\affiliation{
\institution{$^{3}$ Tianjin University, }
\country{China}
}


\renewcommand{\shortauthors}{Xiaoguang Li et al.}

\begin{abstract}
Shadow removal is to restore shadow regions to their shadow-free counterparts while leaving non-shadow regions unchanged.
State-of-the-art shadow removal methods train deep neural networks on collected shadow \& shadow-free image pairs, which are desired to complete two distinct tasks via shared weights, \ie, data restoration for shadow regions and identical mapping for non-shadow regions. 
We find that these two tasks exhibit poor compatibility, and using shared weights for these two tasks could lead to the model being optimized towards only one task instead of both during the training process.
Note that such a key issue is not identified by existing deep learning-based shadow removal methods.
To address this problem, we propose to handle these two tasks separately and leverage the identical mapping results to guide the shadow restoration in an iterative manner. 
Specifically, our method consists of three components: an identical mapping branch (IMB) for non-shadow regions processing, an iterative de-shadow branch (IDB) for shadow regions restoration based on identical results, and a smart aggregation block (SAB).
The IMB aims to reconstruct an image that is identical to the input one, which can benefit the restoration of the non-shadow regions without explicitly distinguishing between shadow and non-shadow regions.
Utilizing the multi-scale features extracted by the IMB, the IDB can effectively transfer information from non-shadow regions to shadow regions progressively, facilitating the process of shadow removal.
The SAB is designed to adaptive integrate features from both IMB and IDB. Moreover, it generates a finely tuned soft shadow mask that guides the process of removing shadows.
Extensive experiments demonstrate our method outperforms all the state-of-the-art shadow removal approaches on the widely used shadow removal datasets.
\end{abstract}




\begin{teaserfigure}
\centering
  \includegraphics[width=1.0\linewidth]{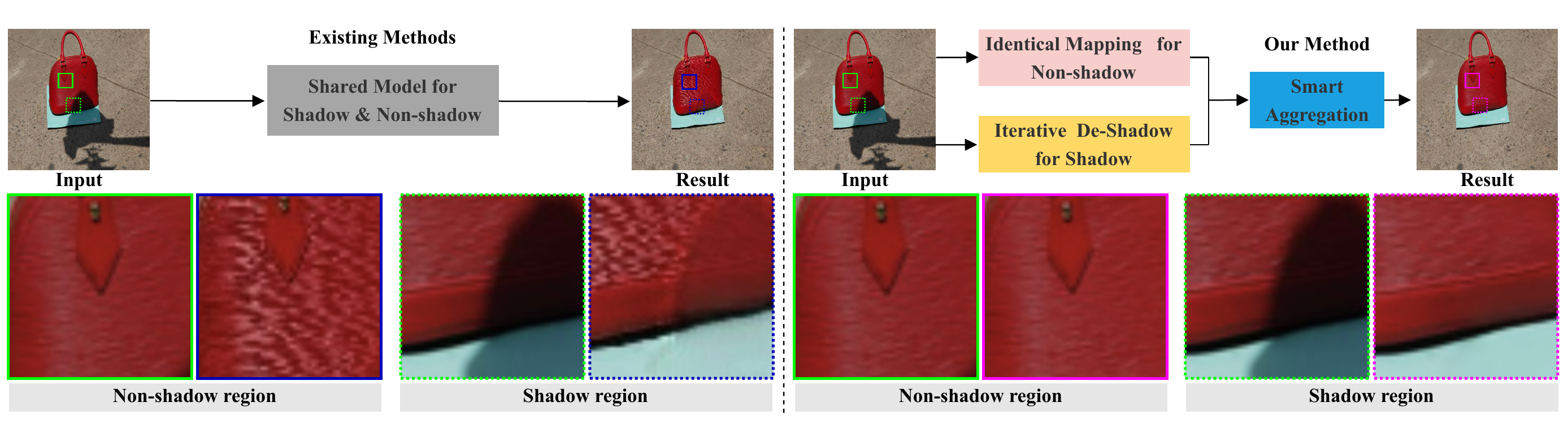}
  \vspace{-25pt}
  \caption{Left: the existing shadow removal methods that use a shared model for restoring both shadow and non-shadow regions. Right: our proposed method that separates shadow removal into two distinct tasks: restoring shadow regions to their shadow-free counterparts and identical mapping for non-shadow regions. The green squares represent the patches in the shadow image. The blue and purple squares represent the patches reconstructed by the shared model and our proposed method, respectively. The solid squares denote the patches in non-shadow regions, and the dashed squares denote the patches in shadow regions.}
  \label{fig:fig1}
\end{teaserfigure}

\maketitle

\section{Introduction}
\label{sec:intro}
The shadows are created when objects block a source of light. Single-image shadow removal aims to reconstruct the shadow-free image from its degraded shadow counterpart, which is an important and non-trivial task in the computer vision field and can benefit many downstream tasks, \eg object detection \cite{nadimi2004physical,fu2021benchmarking, fu2021deep,wang2022yolov7}, objects tracking \cite{sanin2010improved}, and face recognition \cite{zhang2018improving}.
Despite significant progress, shadow removal still faces challenges. Even state-of-the-art shadow removal methods can produce de-shadowed results that contain artifacts in both shadow and non-shadow regions, \eg color inconsistencies between shadow and non-shadow regions, as well as visible marks along the shadow boundary (see \figref{fig:fig1}).

\begin{figure}[t]
\centering
\includegraphics[width=1.0\linewidth]{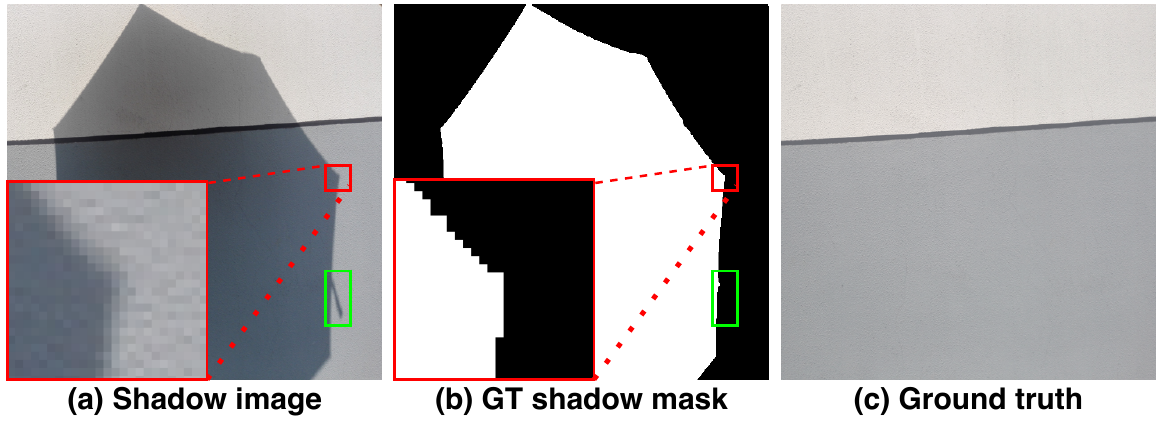}
\vspace{-20pt}
\caption{Visualization of the shadow image and its corresponding ground truth shadow mask. The red square in (a) highlights the gradual transition between the shadow and non-shadow regions, while the red square in (b) denotes the corresponding shadow mask in the same area. The green rectangle in (a) and (b) denote the inconsistency between the shadow regions and the corresponding ground truth shadow mask.
}
\label{fig:introduction_mask}
\vspace{-18pt}
\end{figure}

The existing shadow removal methods often rely on complex deep neural networks to reconstruct both shadow and non-shadow regions simultaneously. However, these methods overlook the fact that shadow removal involves two distinct tasks: restoring shadow regions to their shadow-free counterparts and identical mapping for non-shadow regions. As a consequence, these deep neural networks are optimized towards only one of these tasks during training instead of both, due to the shared weights and the poor compatibility of these tasks. To address this problem, in this work, we propose to handle these two tasks separately. 
Intuitively we can divide shadow removal into these two distinct tasks based on the binary shadow mask. However, due to the diverse properties of shadows in the real world, obtaining an accurate shadow mask that efficiently distinguishes between shadow and non-shadow regions is challenging or impossible. This is particularly true for areas around the shadow boundary, where a gradual transition occurs between shadow and non-shadow regions. Even the ground truth shadow masks provided by the commonly used shadow removal datasets, \eg ISTD+ dataset\cite{le2019shadow}, are not always precise and can not effectively differentiate between shadow and non-shadow regions (see the red and green areas in \figref{fig:introduction_mask}). Therefore, decoupling shadow removal into these two distinct tasks and processing them separately is challenging.

To tackle this issue, we claim that shadow removal can be decoupled by transferring identical mapping without explicitly distinguishing between shadow and non-shadow regions. Specifically, our approach consists of three components: an identical mapping branch (IMB) for processing non-shadow regions, an iterative de-shadow branch (IDB) for shadow regions restoration based on identical results, and a smart aggregation block (SAB). 
The IMB aims to reconstruct an image that is identical to the input one, which can benefit the restoration of the non-shadow regions without explicitly distinguishing between shadow and non-shadow regions.
The IDB is responsible for progressively transferring the information from the non-shadow regions to the shadow regions in an iterative manner to facilitate the process of shadow removal by utilizing the multi-scale features provided by IMB.
The SAB is designed to adaptive integrate features from both IMB and IDB. Moreover, the SAB can generate finely tuned soft shadow masks at multiple feature levels (see \figref{fig:fuse_mask} (c), (d), and (e)) to guide the process of removing shadows. In summary, this work makes the following contributions:

\ding{182} We are the first to decouple the shadow removal problem into two distinct tasks: restoring shadow regions to their shadow-free counterparts and identical mapping for non-shadow regions and propose a novel Dual-Branch shadow removal paradigm for solving this problem.

\ding{183} We propose a novel Dual-Branch shadow removal network that uses an identical mapping branch (IMB) to process the non-shadow regions, an iterative de-shadow branch (IDB) to process the shadow regions, and a smart aggregation block (SAB) to adaptive  aggregate features from two branches.

\ding{184} The extensive experiments demonstrate our proposed method outperforms all previous state-of-the-art shadow removal approaches on several public shadow removal datasets, \ie ISTD+ and SRD.

\section{Related Work}

\subsection{General Shadow Removal Methods}
To restore the shadow-free image from the degraded shadow counterpart, traditional methods \cite{mohan2007editing,finlayson2005removal,guo2012paired,zhang2015shadow,finlayson2009entropy,finlayson2005removal} rely on the priors information, \eg gradients, illumination, and patch similarity.
For example, \cite{mohan2007editing} proposes a gradient-domain processing technique to adjust the softness of the shadows without introducing any artifacts. 
\cite{guo2012paired} uses a region-based approach that predicts relative illumination conditions between segmented regions to distinguish shadow regions and relight each pixel.
%
\cite{zhang2015shadow} extends this region-based approach and constructs a novel illumination recovering operator to effectively remove the shadows and restore the detailed texture information based on the texture similarity between the shadow and non-shadow patches.
Although well-designed, these methods have been surpassed by deep learning-based shadow removal methods recently \cite{wan2022sg-shadow,fu2021auto,zhu2022bijective,inoue2020learning,qu2017deshadownet,le2019shadow,le2020from,hu2019direction,wei2019shadow,zhang2020ris,zhu2022efficient,li2023leveraging,gao2022towards,abiko2022channel}.
%
Specifically, \cite{qu2017deshadownet} is the pioneering work that uses an end-to-end network to tackle shadow detection and shadow removal problems by extracting multi-context features from the global and local regions.
Since then, a large number of interesting deep learning-based methods have been proposed by focusing on different problem aspects.
\cite{fu2021auto} reformulates the shadow removal task as an exposure problem and employs a neural network to predict the exposure parameters to get shadow-free images.
%
\cite{zhu2022bijective} takes into account the auxiliary supervision of shadow generation in the shadow removal procedure and proposes a unified network to perform shadow removal and shadow generation.
%
\cite{wan2022sg-shadow} explicitly considers the style consistency between shadow and non-shadow regions after shadow removal and proposes a style-guided shadow removal network.
%
Although these methods achieve promising results, the problem of lacking large-scale paired training data becomes a bottleneck that limits their performance.
To alleviate this problem, \cite{inoue2020learning} designs a pipeline to generate a large-scale synthetic shadow dataset to improve the shadow removal performance.
While the unsupervised or weekly supervised methods are also introduced \cite{liu2021from,liu2021shadow,hu2019maskshadowgan,jin2021dc,he2021unsupervised} to train the deep neural network with unpaired data.

\subsection{Iterative Network}
Iterative networks are extensively employed in machine learning-based image processing tasks \cite{yan2022rignet,zhou2022edge,li2022practical,hang2019cascaded,saharia2022image,wang2022iterative,zhan2019esir,yu2019deep,li2020recurrent} to recursively and gradually improve the quality of predictions. 
%
For example, 
\cite{yan2022rignet} uses a recurrent image-guided network to address challenges in depth prediction, where the recurrent is applied to both the image guidance branch and the depth generation branch to gradually and sufficiently recover depth values. 
\cite{zhou2022edge} introduces an edge-guided recurrent positioning network for predicting salient objects in optical remote sensing images. The proposed approach can sharpen the predicted positions by utilizing the effective edge information and recurrently calibrating them during the prediction process.
%
%
\cite{hang2019cascaded} introduces a cascaded recurrent neural network that utilizes gated recurrent units to effectively explore the redundant and complementary information present in hyperspectral images.
\cite{saharia2022image} performs image Super-Resolution via Repeated Refinement, which employs a stochastic iterative denoising process to improve image super-resolution performance.
%
%
\cite{zhan2019esir} introduces an end-to-end trainable scene text recognition system that utilizes an iterative rectification framework to address the problem of perspective distortion and text line curvature.
\cite{yu2019deep} introduces a deep iterative down-up convolutional neural network for image denoising that employs a resolution-adaptive approach by iteratively reducing and increasing the resolution at the feature level.
\cite{li2020recurrent} presents a recurrent feature reasoning (RFR) network for single-image inpainting that iteratively predicts hole boundaries at the feature level of a convolutional neural network, which then serves as cues for subsequent inference.
Inspired by the success of these iterative networks, in this paper, we employ an iterative de-shadow branch (IDB) to gradually improve the performance of shadow removal.
%

\section{Discussion and Motivation}
We can divide the shadow removal task into two distinct tasks: 1) restoring shadow regions to their shadow-free counterparts and 2) identical mapping for non-shadow regions.
Previous methods \cite{qu2017deshadownet,fu2021auto,zhu2022bijective} use deep neural networks with shared weights to handle the two tasks. We argue that these networks are only optimized toward one of the tasks instead of both. 
In this section, we first conduct experiments to uncover the limitation of training a shared deep neural network to handle these two distinct tasks (See \secref{mutual_interference}).
%
Then, we illustrate the superiority of identical mapping for non-shadow regions.
To do this, we utilize the Information in the Weights (IIW)\cite{wang2021pac} technique to demonstrate the benefits of restoring non-shadow regions using identical mapping training (See \secref{process_individually}). 
Additionally, we employ an encoder-decoder architecture to explore the efficacy of the iterative technique for shadow removal and highlight its advantages and limitations (See \secref{iterative_network}).

\subsection{Mutual Interference of Shadow Removal Network}
\label{mutual_interference}
\begin{figure}[t]
\centering
\includegraphics[width=1.0\linewidth]{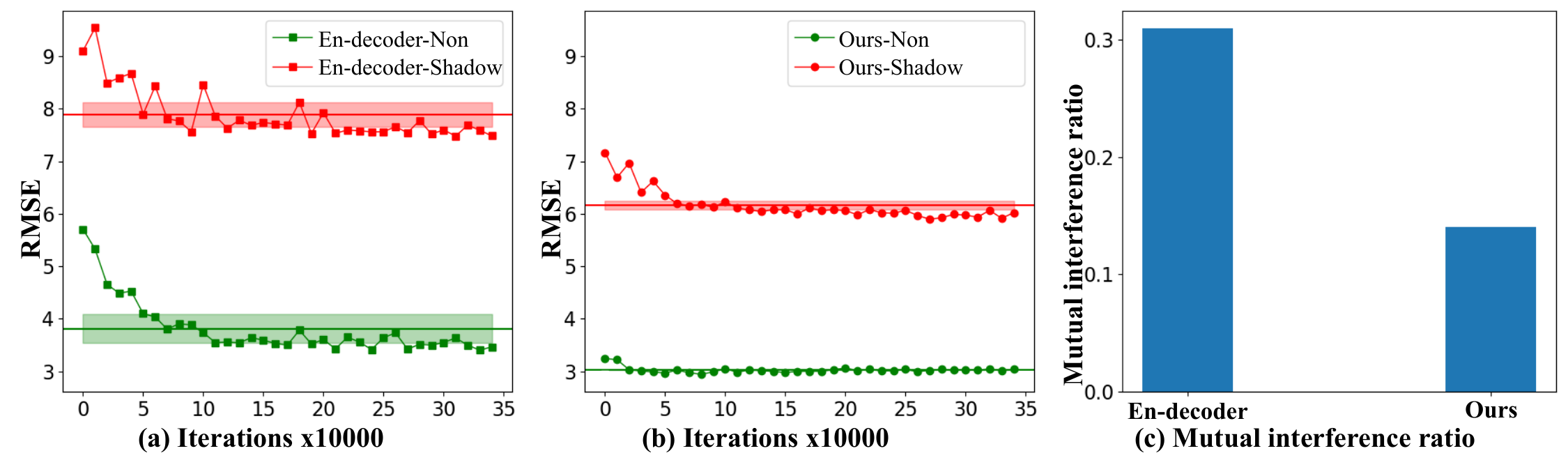}
\vspace{-20pt}
\caption{(a) shows the fluctuation in terms of RMSE during the training in both shadow regions (represented by red squares) and non-shadow regions (represented by green squares) by using the encoder-decoder-based architecture. (b) demonstrate the fluctuation in terms of RMSE during the training in both shadow regions (\ie, red points) and non-shadow regions (\ie, green points) by using our proposed method.
In (a) and (b), the red horizontal lines and the red rectangles indicate the mean and variance of RMSEs in shadow regions, while the green horizontal line and the green rectangles denote the mean and variance of RMSEs in non-shadow regions. (c) presents the mutual interference ratio between shadow and non-shadow restoration.
}
\label{fig:discussion_1}
\vspace{-10pt}
\end{figure}

%
%
To begin with, we define mutual interference as a phenomenon in which the performance of a shadow removal network increases in the shadow restoration task but decreases in the identical mapping task or vice versa. 
%
To uncover this phenomenon, we build a baseline encoder-decoder shadow removal network and evaluate its performance on the ISTD+ dataset\cite{le2019shadow} via root mean squared error (RMSE) every 10000 iterations during training. 
Specifically, given the shadow removal network trained at $t$th iteration, we can calculate the RMSEs at both shadow and non-shadow regions of testing images. Then, we can draw plots along the iteration indexes for both shadow and non-shadow regions (\ie, the red plot and green plot in \figref{fig:discussion_1} (a)). 
Then, we say that mutual interference occurs if the RMSE variation between two neighboring iterations at the shadow region is different from the one at the non-shadow region. 
We define the mutual interference ratio as the number of times mutual interference occurs during the training procedure. 
As depicted in \figref{fig:discussion_1}(c), we see that the baseline encoder-decoder network has a high mutual interference ratio (\ie, over 30 times within 100 counted iterations), which illustrates that a shared deep neural network can hardly cover the two distinct tasks at the same time.

\subsection{Advantages of The Identical Mapping}
\label{process_individually}
\begin{figure}[t]
\centering
\includegraphics[width=1.0\linewidth]{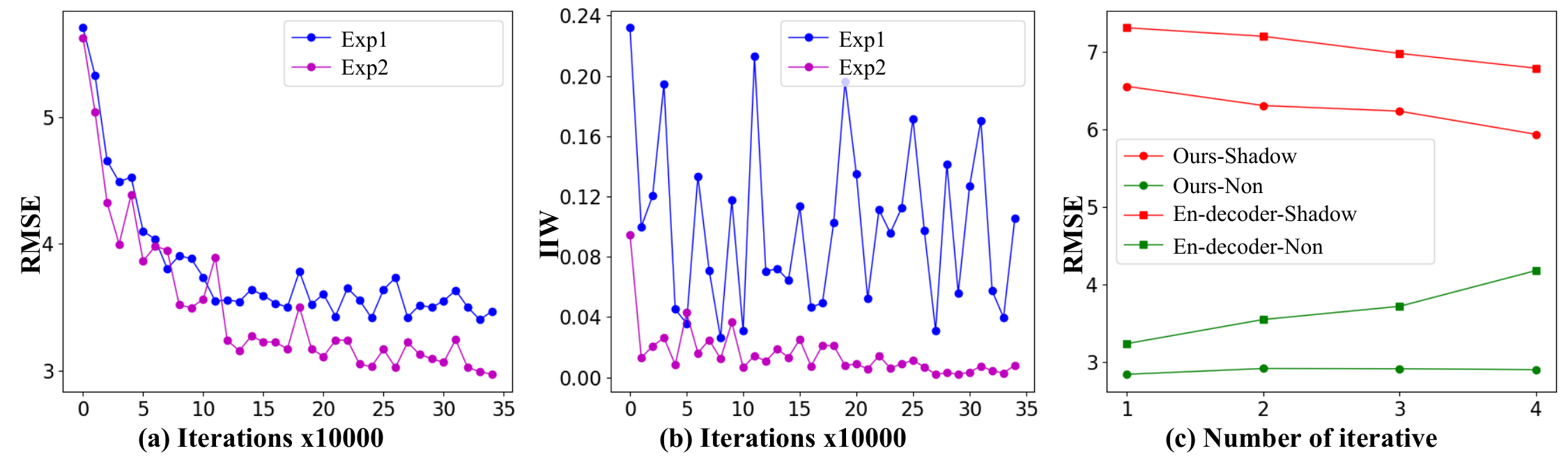}
\vspace{-20pt}
\caption{(a) illustrates the RMSE fluctuation in Exp1 (represented by blue points) and Exp2 (represented by magenta points), respectively, in the non-shadow regions.
(b) illustrates the IIW fluctuation in Exp1 (represented by blue points) and Exp2 (represented by magenta points), respectively, in the non-shadow regions.
In (c), we compare the encoder-decoder structure with our proposed method in both shadow and non-shadow regions using different numbers of iterations.
}
\label{fig:discussion_2}
\vspace{-15pt}
\end{figure}

To analyze the potential advantages of identical mapping, we conduct two additional experiments using the same encoder-decoder network.
%
For Exp1, the input is the shadow image, and the network is trained to remove shadows from that. While for Exp2, we aim to have the reconstructed result identical to the input shadow images (\ie, identical mapping).
For both experiments, we evaluate the restoration quality in the non-shadow regions.
%
%
Surprisingly, we observed that Exp2 had a significant advantage over Exp1. As shown in \figref{fig:discussion_2} (a), in the non-shadow regions, Exp2 has a much lower RMSE than Exp1 throughout the entire training procedure.
To further explore the potential functionality of identical mapping, we used the Information in the Weight (IIW) \cite{wang2021pac} technique to analyze the training procedure of Exp1 and Exp2. 
A lower IIW means the network has a higher generalization to different non-shadow scenes.
The results are displayed in \figref{fig:discussion_2} (b). We observe that Exp2 substantially improved the model's generalization (\ie, lower IIW) in the non-shadow regions compared to Exp1, which demonstrates the efficacy of identical mapping in reconstructing non-shadow regions.
%
%
%

\subsection{Iterative Network for Shadow Removal}
\label{iterative_network}

\begin{figure}[t]
\centering
\includegraphics[width=1.0\linewidth]{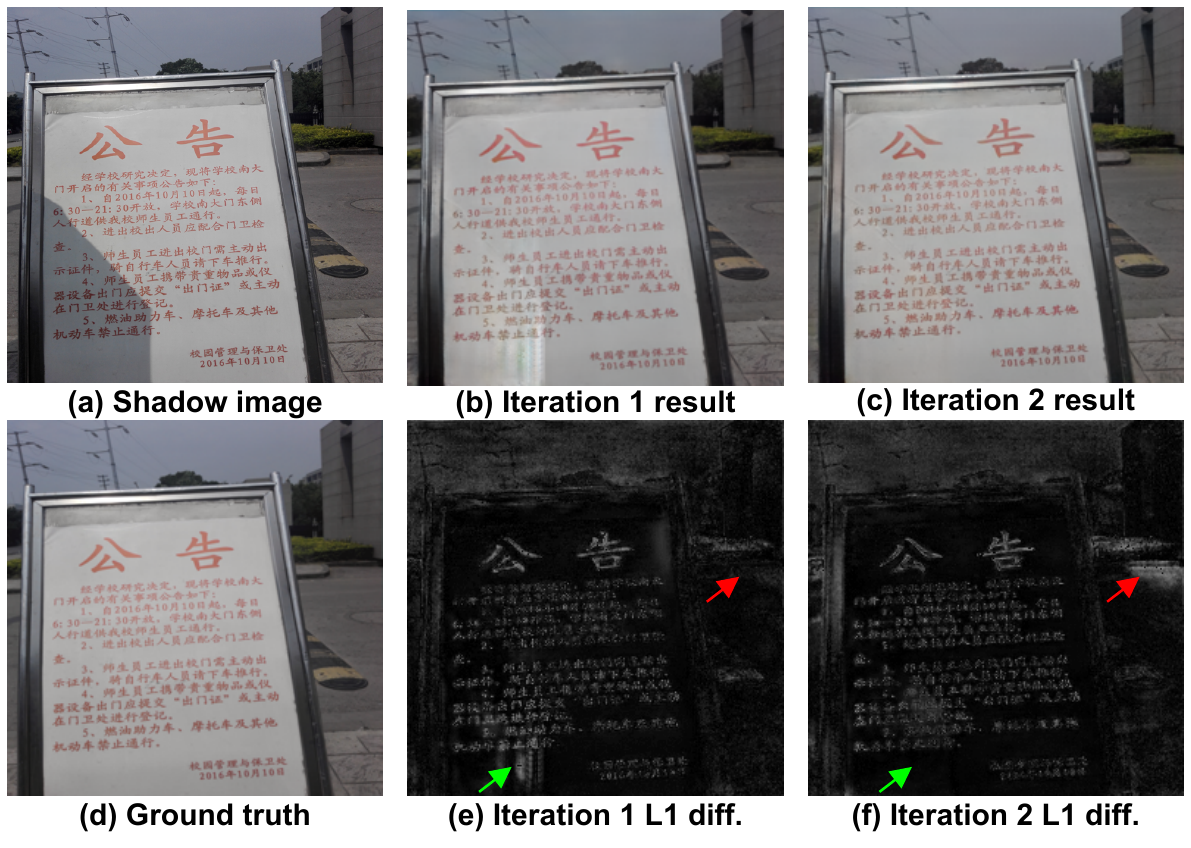}
\vspace{-20pt}
\caption{
Visualized results of the encoder-decoder-based iterative approach by using the number of iterations 1 and 2. (a) and (d) show the shadow image and its corresponding ground truth. (b) and (e) display the reconstruction result with the number of iteration 1 and the L1 difference between the reconstructed image and the ground truth. Similarly, (c) and (f) depict the reconstruction result with the number of iterations 2 and the L1 difference between the reconstructed image and the ground truth.
}
\label{fig:discussion_v}
\vspace{-15pt}
\end{figure}

To investigate the effectiveness of the iterative network, we conduct a series of experiments using the same encoder-decoder structure in \secref{mutual_interference}. 
Specifically, we feed the decoder's feature back into the encoder at various times and evaluate the restoration quality in both shadow and non-shadow regions. 
As depicted in \figref{fig:discussion_2}(c), both the encoder-decoder structure and our proposed method show significant improvement in the restoration quality of shadow regions with more iterations (see the red line).
However, we notice that for the encoder-decoder structure, the restoration quality in non-shadow regions decreases as the number of iterations increases (see the square green line). In contrast, with our method, the restoration quality in non-shadow regions remains nearly unchanged even with increasing iterations (see the dotted green line). 
In addition, we also calculate and visualize the $\mathbf{L}_1$ differences between the ground truth and reconstructed results obtained by the encoder-decoder structure at different iterations. The results are shown in \figref{fig:discussion_v}, where green arrows highlight the differences in shadow regions, and red arrows highlight the differences in non-shadow regions. With a single iteration, the restoration quality is favorable in non-shadow regions but subpar in shadow regions, while the opposite trend is observed with two iterations.

\textit{Overall, the above experiments present the necessity of decoupling the shadow removal task into two distinct tasks and the effectiveness of the combination of identical mapping and iterative shadow removal.}

\section{Method}

\subsection{Overview}
\label{Overview} 
The proposed method consists of three components: an identical mapping branch (IMB) (see \secref{identical_mapping}), an iterative de-shadow branch (IDB) (see \secref{de-shadow-branch}), and a smart aggregation block (SAB) (see \secref{Aggregation_Block}).
Given a shadow image, we decouple the shadow removal into two distinct tasks: restoring shadow regions to their shadow-free counterparts and identical mapping for non-shadow regions. We use IMB to handle the identical mapping task and use IDB to handle the shadow restoration task. The SAB is designed to adaptive integrate features from both IMB and IDB. 
To prove the advantage of our method, we conduct the same experiment as discussed in \secref{mutual_interference} by using our method. As shown in \figref{fig:discussion_1}(b), for both tasks, our method exhibits less fluctuation in terms of RMSE compared to the encoder-decoder structure during training. Besides, our method also demonstrates a lower mutual interference ratio, as shown in \figref{fig:discussion_1}(c).
%
\begin{figure*}[t]
\centering
\includegraphics[width=1.0\linewidth]{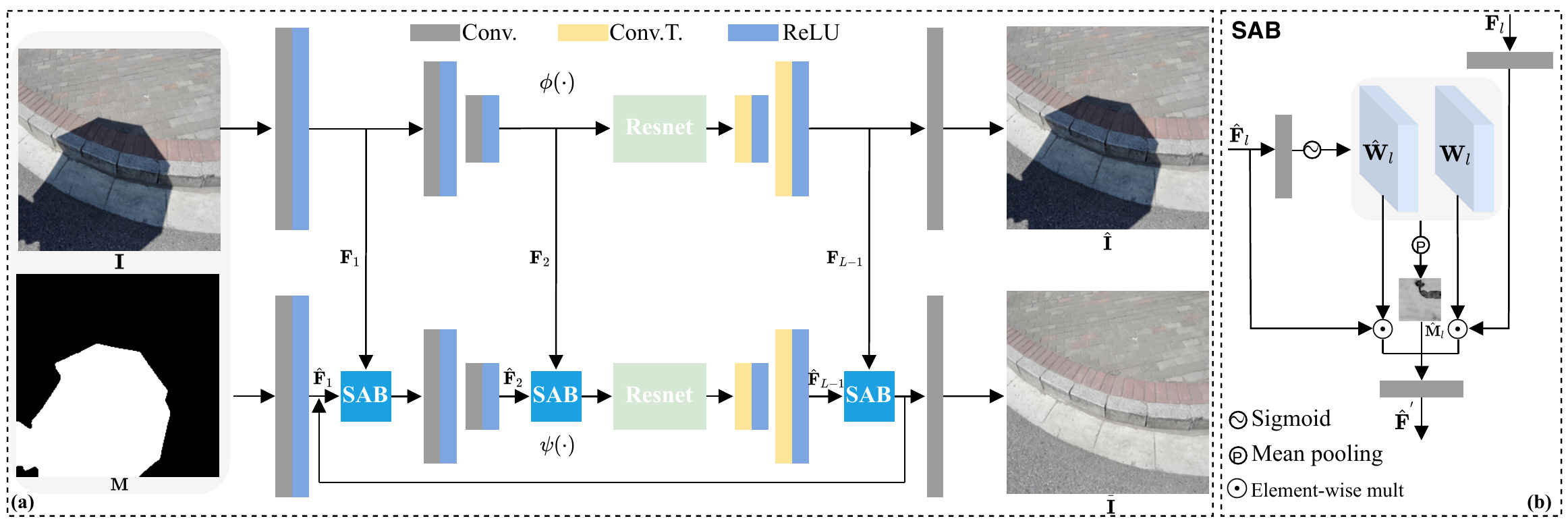}
\vspace{-20pt}
\caption{
(a) and (b) show the architecture of our proposed method and the smart aggregation block, respectively. In (a), the top section represents the identical mapping branch (IMB), and the bottom section represents the iterative de-shadow branch (IDB). The symbol $\mathbf{L}$ denotes the number of convolution layers in the IDB.
}
\label{fig:frameworks}
\vspace{-5pt}
\end{figure*}

\subsection{Identical Mapping Branch}
\label{identical_mapping} 
We propose the identical mapping branch (IMB) to reconstruct the input image via an encoder-decoder network, which can benefit the restoration of non-shadow regions. 
%
Given a shadow image $\mathbf{I}$, the objective of IMB is to reconstruct $\hat{\mathbf{I}}$, where $\hat{\mathbf{I}}$ should be identical to $\mathbf{I}$. This procedure can be represented by
\begin{align}
\label{eq:IMB_1}
\hat{\mathbf{I}} = \phi(\mathbf{I}),
\end{align}
where $\phi(\cdot)$ denote the IMB.
Let $\mathbf{F}_l$ denote the feature extracted from the $l$th convolution layer of $\phi(\cdot)$. $\mathbf{F}_l$ can be formalized as
\begin{align}
\label{eq:IMB_2}
\mathbf{F}_l = \phi_l(\ldots\phi_2(\phi_1(\mathbf{I}))),
\end{align}
where $\phi_l(\cdot)$ denotes the $l$th convolution layer of $\phi(\cdot)$. After training the IMB, we freeze its parameters and rely solely on the multi-scale features, \ie $\mathbf{F}_l$, to guide the iterative de-shadow branch (IDB).
\subsection{Iterative De-shadow Branch}
\label{de-shadow-branch}
The iterative de-shadow branch (IDB) is responsible for progressively transferring the information from the non-shadow regions to the shadow regions in an iterative manner to facilitate the process of shadow removal by utilizing the multi-scale features provided by IMB.
%
%
%
Let $\psi(\cdot)$ denotes the IDB and $\hat{\mathbf{F}}_l$ denote the feature extracted from the $l$th convolution layer of $\psi(\cdot)$. $\hat{\mathbf{F}}_l$ can be formalized as
\begin{align}
\label{eq:DIDB_2}
\hat{\mathbf{F}}_l = \psi_l(\ldots\psi_2(\psi_1(\mathbf{Cat}(\mathbf{I}, \mathbf{M}))))
\end{align}
where $\mathbf{Cat}$ means channel-wise concatenation, $\psi_l(\cdot)$ denotes the $l$th convolution layer of $\psi(\cdot)$ and $\mathbf{M}$ denotes the corresponding binary shadow mask of the shadow image $\mathbf{I}$. The shadow and non-shadow regions are annotated by $\mathbf{1}$ and $\mathbf{0}$ respectively in $\mathbf{M}$.
Then we aggregate $\mathbf{F}_l$ and $\hat{\mathbf{F}}_l$ in an adaptive manner at the multi-scale features level (\ie after the first, third, and second-to-last convolution layers of $\mathbf{\psi(\cdot)}$ as illustrated in \figref{fig:frameworks}). The procedure of the aggregation can be represented as
\begin{align}
\label{eq:DIDB_3}
\hat{\mathbf{F}}_l^{'} = \mathbf{SAB}(\mathbf{F}_l, \hat{\mathbf{F}}_l),
\end{align}
where $\hat{\mathbf{F}}_l^{'}$ denotes the adaptive aggregated feature which will be used as the input of the next convolution layer of $\psi(\cdot)$, \ie $\psi_{l+1}(\cdot)$, and $\mathbf{SAB}$ denotes the smart aggregation block (see \secref{Aggregation_Block}).

\subsection{Smart Aggregation Block}
\label{Aggregation_Block}
\begin{figure}[t]
\centering
\includegraphics[width=1.0\linewidth]{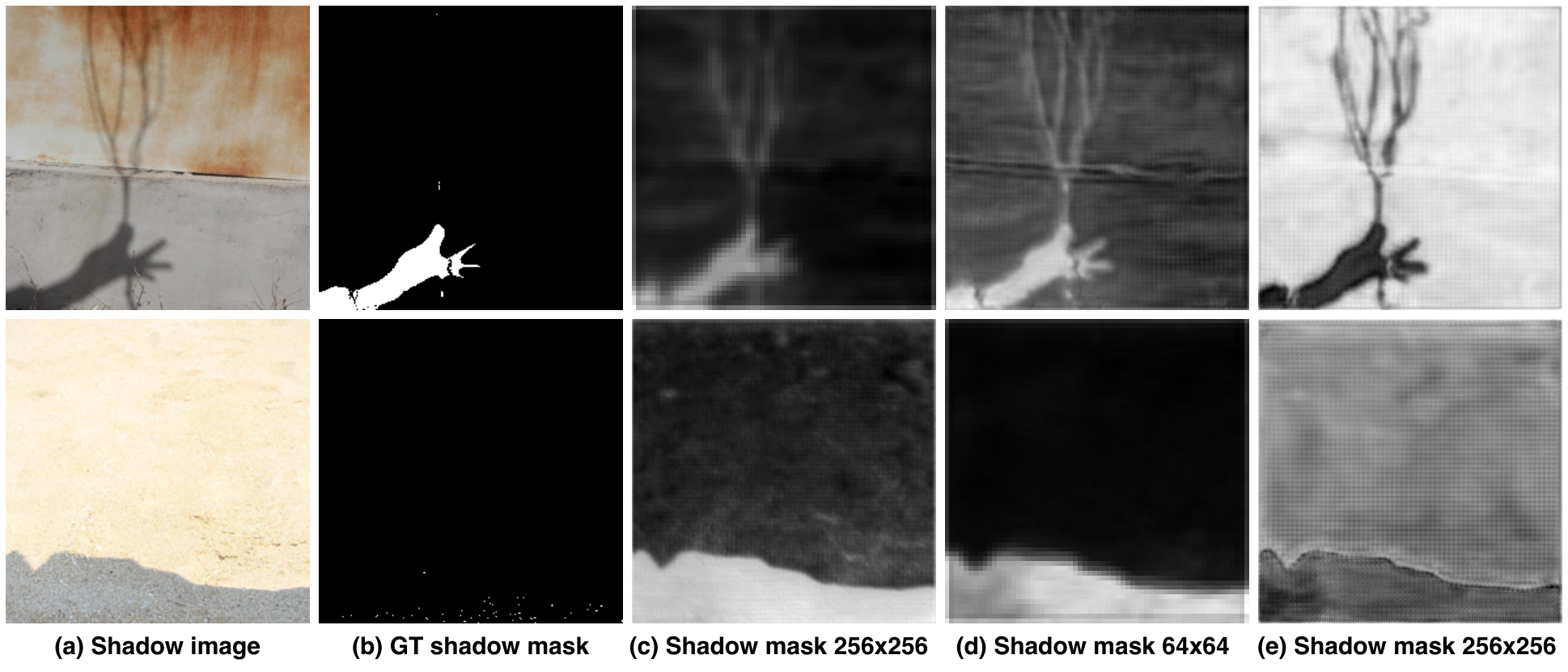}
\vspace{-20pt}
\caption{
Visualized results of the ground truth shadow mask and soft mask produced by the smart aggregation block: (a) and (b) denote the shadow image and ground truth shadow mask, while (c), (d), and (e) show the soft shadow masks produced by the smart aggregation block after the first, third, and second-to-last layers of $\mathbf{\psi(\cdot)}$, respectively.
}
\label{fig:fuse_mask}
\vspace{-15pt}
\end{figure}

Instead of directly concatenating the features extracted from the IMB and IDB, we propose to aggregate them in an adaptive manner. Specifically, we utilize a convolutional layer with a kernel size of 3x3 followed by a sigmoid activation function to estimate the adaptive aggregation weights, which can be represented as
\begin{align}\label{eq:w1_w2}
   [\mathbf{W}_l,\hat{\mathbf{W}}_l]  = \text{Sigmoid(Conv}_\text{weight}(\hat{\mathbf{F}}_l)),
\end{align}
where $\mathbf{W}_l,\hat{\mathbf{W}}_l$ denote the corresponding aggregation weights of $\mathbf{F}_l$, $\hat{\mathbf{F}}_l$ respectively. 
Since the IMB is frozen, $\mathbf{F}_l$ remains constant throughout the iterative procedure of IDB. Therefore, we utilize only $\hat{\mathbf{F}}_l$ to predict the aggregation weights. The whole aggregation procedure can be formalized as
\begin{align}\label{eq:SAB_2}
   \hat{\mathbf{F}}^{'}_l = \mathbf{SAB}(\mathbf{F}_l, \hat{\mathbf{F}}_l) = \text{Conv}_\text{agg.}(
   \mathbf{Cat}(\mathbf{F}_l\odot\mathbf{W}_l, \hat{\mathbf{F}}_l\odot\hat{\mathbf{W}}_l, \hat{\mathbf{M}}_l)),
\end{align}
where $\odot$ denotes the element-wise multiplication, $\text{Conv}_\text{agg.}$ denotes a convolution layer with a kernel size of 3x3, $\hat{\mathbf{M}}_l$ denotes the generated soft shadow mask which is obtained by applying average pooling along the channels of $ [\mathbf{W}_l,\hat{\mathbf{W}}_l]$. 
As shown in \figref{fig:fuse_mask}, the soft shadow masks (see (c)-(e)) obtained from the procedure of aggregation are capable of accurately capturing the shadow regions, in contrast to the binary shadow mask (see (b)) provided by the shadow removal dataset, \ie SRD dataset\cite{qu2017deshadownet}, which can not capture the shadow details, especially for the regions along the shadow boundary.
%


\subsection{Implementation Details}
\begin{table*}[t]
\setlength{\tabcolsep}{4pt}
\footnotesize
\centering
\caption{Detailed architecture of our method. The output size of each layer is defined as H $\times$ W. The parameters of 'Conv\&ConvTran' are the numbers of input and output channels, kernel size, stride, and padding, respectively. $\mathbf{L}$ denotes the number of convolution layers.}
\vspace{-10pt}
\small
{
    \resizebox{0.9\linewidth}{!}{
\begin{tabular}{l|l|l|l||l|l|l|l}
\toprule
\multicolumn{4}{c||}{DIDB($\psi(\cdot)$)} & \multicolumn{4}{c}{IMB ($\phi(\cdot)$)}  \\
\midrule
Input & Output & Output size & Operation & Input & Output & Output size & Operation                 \\
$[\mathbf{I}, \mathbf{M}]$  & $\hat{\mathbf{F}}_1$  & $256\times256$    & Conv(4, 64, 7, 1, 3), ReLU    & 
$\mathbf{I}$              & $\mathbf{F}_1$        & $256\times256$    & Conv(3, 64, 7, 1, 3), ReLU 
\\
$[\mathbf{F}_1, \hat{\mathbf{F}}_1]$ & $\hat{\mathbf{F}}_1^{'}$ & $256\times256$ & SAB                                 &                      &                &                         &   
\\
$\hat{\mathbf{F}}_1^{'}$  & $\hat{\mathbf{F}}_2$  & $128\times128$    & Conv(64, 128, 4, 2, 1), ReLU    & 
$\mathbf{F}_1$              & $\mathbf{F}_2$        & $128\times128$    & Conv(64, 128, 4, 2, 1), ReLU    
\\
$\hat{\mathbf{F}}_2$      & $\hat{\mathbf{F}}_3$  & $64\times64$    & Conv(128, 256, 4, 2, 1), ReLU    & 
$\mathbf{F}_2$              & $\mathbf{F}_3$        & $64\times64$    & Conv(128, 256, 4, 2, 1), ReLU    
\\
$[\mathbf{F}_3, \hat{\mathbf{F}}_3]$ & $\hat{\mathbf{F}}_3^{'}$ & $64\times64$ & SAB                                 &                      &                &                         &   
\\
\midrule
\multicolumn{4}{c||}{Resnet x 8} & \multicolumn{4}{c}{Resnet x 8}  \\
$\hat{\mathbf{F}}_3^{'}$  & $\hat{\mathbf{F}}_4$  & $64\times64$    & Conv(256, 256, 3, 1, 1), ReLU    & 
$\mathbf{F}_3$            & $\mathbf{F}_4$        & $64\times64$    & Conv(256, 256, 3, 1, 1), ReLU    
\\
$\hat{\mathbf{F}}_4$      & $\hat{\mathbf{F}}_4$  & $64\times64$    & Conv(256, 256, 3, 1, 1), ReLU    & 
$\mathbf{F}_4$            & $\mathbf{F}_4$        & $64\times64$    & Conv(256, 256, 3, 1, 1), ReLU    
\\
\midrule
$\hat{\mathbf{F}}_{L-3}$  & $\hat{\mathbf{F}}_{L-2}$  & $128\times128$    & Conv(256, 128, 4, 2, 1), ReLU    & 
$\mathbf{F}_{L-3}$               & $\mathbf{F}_{L-2}$        & $128\times128$    & Conv(256, 128, 4, 2, 1), ReLU    
\\
$\hat{\mathbf{F}}_{L-2}$  & $\hat{\mathbf{F}}_{L-1}$  & $256\times256$    & Conv(128, 64, 4, 2, 1), ReLU    & 
$\mathbf{F}_{L-2}$              & $\mathbf{F}_{L-1}$        & $256\times256$    & Conv(128, 64, 4, 2, 1), ReLU    
\\
$[\mathbf{F}_{L-1}, \hat{\mathbf{F}}_{L-1}]$ & $\hat{\mathbf{F}}_{L-1}^{'}$ & $256\times256$ & SAB                                 &                      &                &                         & 
\\
$\hat{\mathbf{F}}_{L-1}^{'}$  & $\hat{\mathbf{F}}_L$  & $256\times256$    & Conv(64, 3, 7, 1, 3)   
    &                      &                &                         &     
\\
\bottomrule
\end{tabular}
}
}
\vspace{-10pt}
\label{tab:network_detail}
\end{table*}

{\bf Network architectures.} Following \cite{li2022misf, guo2021jpgnet} the identical mapping branch $\mathbf{\phi(\cdot)}$ and the iterative de-shadow branch $\mathbf{\psi(\cdot)}$ employ a similar encoder-decoder architecture with different input and output, as shown in \tableref{tab:network_detail}. 
Theoretically, the smart aggregation block can be added after each convolution layer of $\mathbf{\psi(\cdot)}$ to maximize its potential impact. However, to optimize computation efficiency, in our experiments, we selectively add the smart aggregation block after the first, third, and second-to-last layers of $\mathbf{\psi(\cdot)}$ to balance the computation and performance.

{\bf Loss functions.} Following the previous shadow removal method \cite{fu2021auto}, we only employ the $\mathbf{L}_1$ loss during the training process. Specifically, we first train the identical Mapping branch $\phi(\cdot)$ with the objective function
\begin{align}\label{eq:loss_1}
\mathcal{L}_1(\hat{\mathbf{I}}, \mathbf{I}) = \|\hat{\mathbf{I}}-\mathbf{I}\|_1.
\end{align}
Then we freeze $\phi(\cdot)$ and train the iterative de-shadow branch $\psi(\cdot)$ with the same objective function
\begin{align}\label{eq:loss_2}
\mathcal{L}_2(\mathbf{I}^*, \overline{\mathbf{I}}) = \|\mathbf{I}^*-\overline{\mathbf{I}}\|_1,
\end{align}
where $\overline{\mathbf{I}}$ and $\mathbf{I}^*$ denote the de-shadowed result and the corresponding ground truth shadow-free image, respectively. 

{\bf Training details.} We adopt a two-step training strategy. Firstly, we exclusively utilize shadow images to train the identical mapping branch $\phi(\cdot)$ for 500,000 iterations, employing a batch size of 8. Subsequently, we freeze $\phi(\cdot)$ and utilize paired shadow \& shadow-free images to train the iterative de-shadow branch $\psi(\cdot)$ for 150,000 iterations with the same batch size. Following \cite{fu2021auto}, we resize the input shadow image to 256x256 resolution. Both branches are optimized using the Adam optimizer with a learning rate of 0.00005. All experiments are conducted on the Linux server equipped with two NVIDIA Tesla V100 GPUs.

\section{Experiments}

\subsection{Setups}

{\bf Datasets.}
Following the previous shadow removal method \cite{wan2022sg-shadow}, we conduct experiments on two widely used shadow removal datasets \ie ISTD+ \cite{le2019shadow} and SRD \cite{qu2017deshadownet}.
%
%
The ISTD+ dataset consists of 1330 triplets for training and 540 triplets for testing. We use the provided ground truth masks directly during the training procedure. While in the evaluation step, we follow the previous method \cite{fu2021auto} and use Ostu’s algorithm to detect the corresponding shadow masks.
The SRD dataset contains 2680 paired shadow and shadow-free images for training and 408 paired shadow and shadow-free images for testing. Because the SRD does not provide the corresponding shadow masks, we use the shadow masks provided by DHAN\cite{cun2020towards} for both training and evaluation steps.

{\bf Metrics.}
We adopt a comprehensive evaluation approach for assessing the performance of our proposed method. Firstly, we calculate the root mean square error (RMSE) in the LAB color space. Furthermore, following the previous approaches \cite{zhu2022bijective, wan2022sg-shadow}, we employ the commonly used image quality evaluation metrics, \ie peak signal-to-noise ratio (PSNR) \cite{johnson2016perceptual}, structural similarity index (SSIM), and learned perceptual image patch similarity (LPIPs) \cite{zhang2018unreasonable}. This allows us to thoroughly evaluate the restoration quality of our proposed method from multiple perspectives.

{\bf Baselines.}
We conduct comprehensive comparisons with previous state-of-the-art shadow removal algorithms, including SP+M-Net \cite{le2019shadow}, Param+M+D-Net \cite{le2020from}, Fu et al. \cite{fu2021auto}, LG-ShadowNet \cite{liu2021shadow}, DC-ShadowNet \cite{jin2021dc}, G2R-ShadowNet \cite{liu2021from}, BMNet \cite{zhu2022bijective}, and SG-ShadowNet \cite{wan2022sg-shadow} on the ISTD+ dataset. 
Additionally, we compare with DSC \cite{hu2019direction}, DHAN \cite{cun2020towards}, Fu et al. \cite{fu2021auto}, DC-ShadowNet \cite{jin2021dc}, BMNet \cite{zhu2022bijective}, and SG-ShadowNet \cite{wan2022sg-shadow} on the SRD dataset.

\subsection{Comparison Results}
{\bf Quantitative comparison.} 
To validate the effectiveness of our proposed method, we first conduct a comprehensive comparison with recent state-of-the-art methods on the ISTD+ datasets. The results, as depicted in \tableref{tab:comparison_ISTD+}, clearly demonstrate our method's superiority in terms of reconstruction quality, as evaluated by multiple metrics, including RMSE, PSNR, SSIM, and LPIPS. 
Specifically, for the comparison at the whole image level, our method outperforms all the competitors. Compared to Fu et al.\cite{fu2021auto}, our method achieves a reduction of 20.50\% in RMSE and 68.38\% in LPIPS, as well as an increase of 15.32\% in PSNR and 14.21\% in SSIM.
Similarly, compared to Param+M+D-Net\cite{le2020from}, our method demonstrates superior performance with a reduction of 15.92\% in RMSE and 30.30\% in LPIPS, as well as an increase of 12.68\% in PSNR and 1.89\% in SSIM.
For the comparison in the shadow regions, our method also outperforms other methods. Compared to Fu et al.\cite{fu2021auto}, our method achieves a reduction of 9.80\% in RMSE, as well as an increase of 5.14\% in PSNR and 1.30\% in SSIM.
When compared to Param+M+D-Net\cite{le2020from}, our method achieves a reduction of 38.87\% in RMSE, as well as an increase of 13.96\% in PSNR and 0.47\% in SSIM.
For the comparison in the non-shadow regions, our method continues to outperform other methods. Compared to Fu et al.\cite{fu2021auto}, our method achieves a reduction of 24.12\% in RMSE, as well as an increase of 19.45\% in PSNR and 11.54\% in SSIM. 
Additionally, when compared to Param+M+D-Net\cite{le2020from}, our method achieves a reduction of 1.06\% in RMSE, as well as an increase of 7.89\% in PSNR and 0.43\% in SSIM.

To further substantiate the effectiveness of our proposed method, we conducted additional comparison experiments on the SRD dataset. The results, as presented in \tableref{tab:comparison_SRD}, demonstrate the superiority of our method over other state-of-the-art shadow removal approaches. Our method exhibits a significant margin of improvement across all evaluation metrics.
Specifically, for the comparison at the whole image level, our method outperforms all the competitors. Compared to DHAN\cite{cun2020towards}, our method achieves a reduction of 22.20\% in RMSE and 9.09\% in LPIPS, as well as an increase of 9.46\% in PSNR and 1.47\% in SSIM.
Compared to BMNet\cite{zhu2022bijective}, our method demonstrates superior performance with a reduction of 14.39\% in RMSE and 11.87\% in LPIPS, as well as an increase of 5.30\% in PSNR and 0.41\% in SSIM.
For the comparison in the shadow regions, our method also outperforms other methods. Compared to DHAN\cite{cun2020towards}, our method achieves a reduction of 24.44\% in RMSE, as well as an increase of 7.15\% in PSNR and 0.41\% in SSIM.
When compared to BMNet\cite{zhu2022bijective}, our method achieves a reduction of 15.90\% in RMSE, as well as an increase of 6.12\% in PSNR and 0.43\% in SSIM.
For the comparison in the non-shadow regions, our method continues to outperform other methods. Compared to DHAN\cite{cun2020towards}, our method achieves a reduction of 20.31\% in RMSE, as well as an increase of 9.22\% in PSNR and 0.93\% in SSIM. 
Additionally, when compared to BMNet\cite{zhu2022bijective}, our method achieves a reduction of 13.13\% in RMSE, as well as an increase of 2.65\% in PSNR and 0.04\% in SSIM.
These comparison results unequivocally support the effectiveness of our proposed method and its superiority over the state-of-the-art methods in relation to reconstruction quality in both the shadow regions and non-shadow regions.

\begin{table*}[t]

\footnotesize
\centering
\caption{Comparison results on ISTD+ dataset\cite{le2019shadow}.}
\vspace{-10pt}
\small
{
    \resizebox{0.9 \linewidth}{!}{
    {
	\begin{tabular}{l|cccc|ccc|ccc}
		
    \toprule
    
     \multirow{2}{*}{Method} 
     & \multicolumn{4}{c|}{All} 
     & \multicolumn{3}{c|}{Shadow} 
     & \multicolumn{3}{c}{Non-Shadow}   \\

    & RMSE$\downarrow$ & PSNR$\uparrow$ & SSIM$\uparrow$ & LPIPS$\downarrow$
    & RMSE$\downarrow$ & PSNR$\uparrow$ & SSIM$\uparrow$
    & RMSE$\downarrow$ & PSNR$\uparrow$ & SSIM$\uparrow$ \\
    
    \midrule
    
    SP+M-Net\cite{le2019shadow}
    & 3.610 & 32.33 & 0.9479 & 0.0716
    & 7.205 & 36.16 & 0.9871 
    & 2.913 & 35.84 & 0.9723\\

    Param+M+D-Net\cite{le2020from} 
    & 4.045	& 30.12	& 0.9420 & 0.0759
    & 9.714 & 33.59 & 0.9850 
    & 2.935 & 34.33 & 0.9723
     \\

    Fu et al.\cite{fu2021auto}
    & 4.278 & 29.43 & 0.8404 & 0.1673
    & 6.583 & 36.41 & 0.9769
    & 3.827 & 31.01 & 0.8755
     \\

    LG-ShadowNet\cite{liu2021shadow} 
    & 4.402	& 29.20	& 0.9335 & 0.0920
    & 9.709	& 32.65 & 0.9806 
    & 3.363	& 33.36	& 0.9683
     \\

    DC-ShadowNet\cite{jin2021dc} 
    & 4.781	 & 28.76 & 0.9219 & 0.1112
    & 10.434 & 32.20 & 0.9758
    & 3.674  & 33.21 & 0.9630
     \\
    
    G2R-ShadowNet\cite{liu2021from} 
    & 3.970 & 30.49	& 0.9330 & 0.0868
    & 8.872 & 34.01 & 0.9770
    & 3.010 & 34.62 & 0.9707
     \\
    
    BMNet\cite{zhu2022bijective}
    & 3.595 & 32.30 & 0.9551 & 0.0567
    & 6.189 & 37.30 & \topone{0.9899}
    & 3.087 & 35.06 & 0.9738
    \\

    SG-ShadowNet\cite{wan2022sg-shadow}
    & 3.531 & 32.41 & 0.9524 & 0.0594
    & 6.019 & 37.41 & 0.9893
    & 3.044 & 34.95 & 0.9725
     \\

    Ours
    & \topone{3.401} & \topone{33.94} & \topone{0.9598} & \topone{0.0529}
    & \topone{5.938} & \topone{38.28} & 0.9896
    & \topone{2.904} & \topone{37.04} & \topone{0.9765}
     \\
     						
    \bottomrule

	\end{tabular}
	}
	}
}
\label{tab:comparison_ISTD+}
\vspace{0pt}
\end{table*}

\begin{figure*}[t]
\centering
\includegraphics[width=1.0\linewidth]{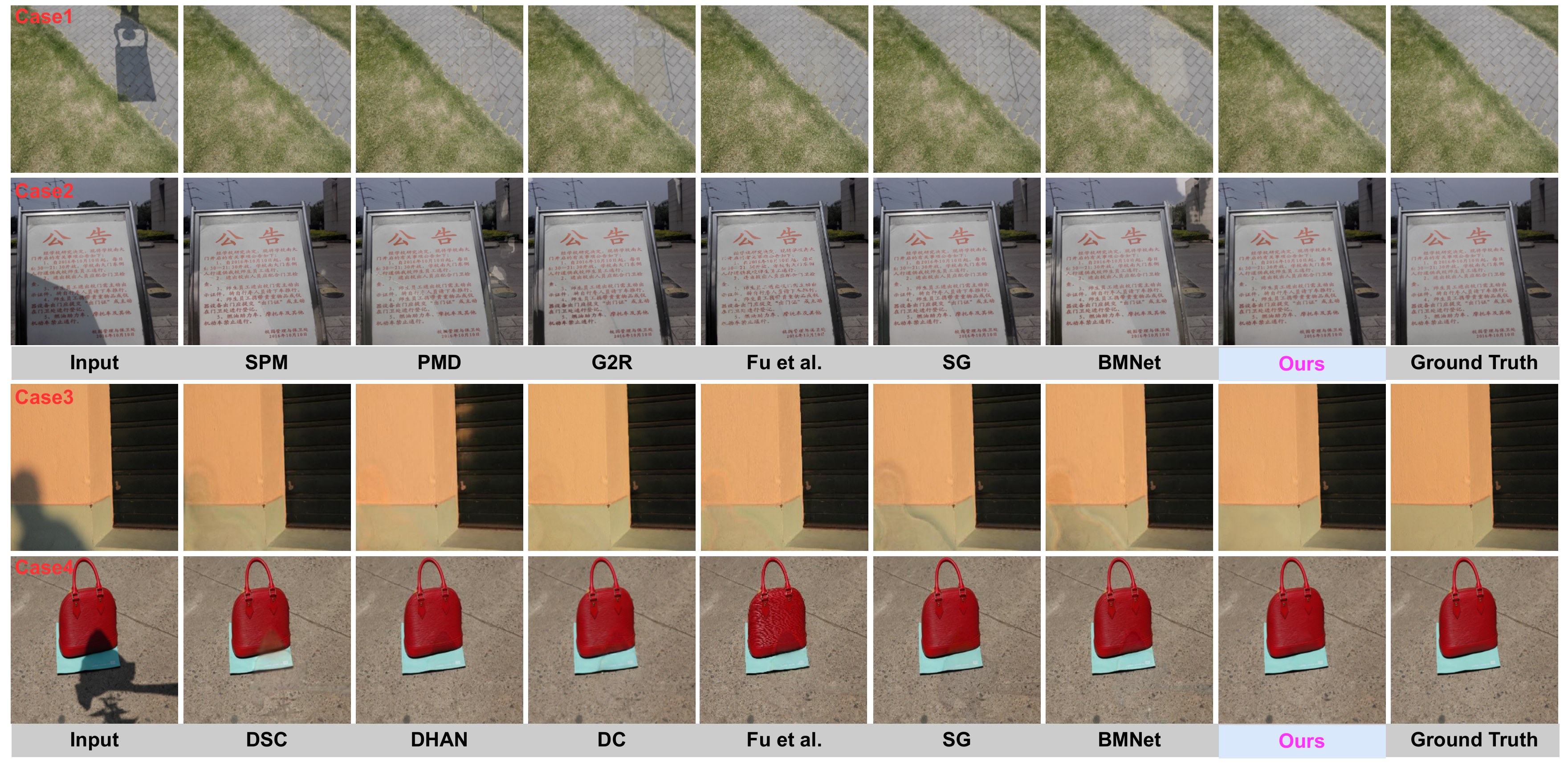}
\vspace{-20pt}
\caption{
Visualization results: the first two rows are from the ISTD+ dataset, and the last two rows are from the SRD dataset.
}
\label{fig:compare_v}
\vspace{-5pt}
\end{figure*}

\begin{table*}[t]
\footnotesize
\centering
\caption{Comparison results on SRD dataset\cite{qu2017deshadownet}.}
\vspace{-10pt}
\small
{
    \resizebox{0.9\linewidth}{!}{
    {
	\begin{tabular}{l|cccc|ccc|ccc}
		
    \toprule
    
     \multirow{2}{*}{Method} 
     & \multicolumn{4}{c|}{All} 
     & \multicolumn{3}{c|}{Shadow} 
     & \multicolumn{3}{c}{Non-Shadow}   \\

    & RMSE$\downarrow$ & PSNR$\uparrow$ & SSIM$\uparrow$ & LPIPS$\downarrow$
    & RMSE$\downarrow$ & PSNR$\uparrow$ & SSIM$\uparrow$ 
    & RMSE$\downarrow$ & PSNR$\uparrow$ & SSIM$\uparrow$ 
    \\
    
    \midrule
    
    DSC\cite{hu2019direction} 
    & 5.704	& 29.01	& 0.9044 & 0.1145
    & 8.828	& 34.20 & 0.9702
    & 4.509	& 31.85	& 0.9555
     \\

    DHAN\cite{cun2020towards} 
    & 4.666	& 30.67	& 0.9278 & 0.0792
    & 7.771 & 37.05 & 0.9818 
    & 3.486	& 32.98	& 0.9591
     \\

    Fu et al.\cite{fu2021auto} 
    & 6.269	& 27.90	& 0.8430 & 0.1820
    & 8.927	& 36.13 & 0.9742
    & 5.259	& 29.43	& 0.8888
     \\

    DC-ShadowNet\cite{jin2021dc} 
    & 4.893	& 30.75	& 0.9118 & 0.1084
    & 8.103	& 36.68 & 0.9759
    & 3.674	& 33.10	& 0.9540
     \\
    
    BMNet\cite{zhu2022bijective}
    & 4.240	& 31.88	& 0.9376 & 0.0817
    & 6.982	& 37.41	& 0.9816
    & 3.198	& 35.09	& 0.9676
     \\

    SG-ShadowNet\cite{wan2022sg-shadow} 
    & 4.297	& 31.31	& 0.9273 & 0.0835
    & 7.564 & 36.55 & 0.9807
    & 3.056	& 34.23	& 0.9611
     \\
    
    Ours 
   				 			
    & \topone{3.630} & \topone{33.57} & \topone{0.9414} & \topone{0.0720}
    & \topone{5.872} & \topone{39.70} & \topone{0.9858}
    & \topone{2.778} & \topone{36.02} & \topone{0.9680}
     \\
    		
    \bottomrule

	\end{tabular}
	}
	}
}
\label{tab:comparison_SRD}
\vspace{-10pt}
\end{table*}
%
{\bf Qualitative comparison.} 
We compared our visualized results with other state-of-the-art shadow removal methods on both ISTD+ and SRD datasets. As shown in \figref{fig:compare_v}, our method consistently outperforms the competitors in two aspects: \ding{182} Our reconstructed results exhibit superior color consistency. In particular, for case 2 and case 4, our method produces color-consistent results where the shadow regions and non-shadow regions are nearly indistinguishable from the human eye. In contrast, the competitors' results show obvious color inconsistency. \ding{183} The mask boundary in our reconstructed results is smoother and more seamless. For case 1, the mask boundary in the competitors' results is clearly visible, whereas it is imperceptible in our method. For case 3, the competitors' results exhibit ghosting artifacts around the mask boundary, while our result does not show any artifacts around the mask boundary. 

\subsection{Ablation Study}

\begin{table*}[t]
\footnotesize
\centering
\caption{Ablation study on ISTD+ dataset\cite{le2019shadow}.}
\vspace{-10pt}
\small
{
    \resizebox{0.9\linewidth}{!}{
    {
	\begin{tabular}{l|cccc|ccc|ccc}
		
    \toprule
     
     \multirow{2}{*}{Method} 
     & \multicolumn{4}{c|}{All} 
     & \multicolumn{3}{c|}{Shadow} 
     & \multicolumn{3}{c}{Non-Shadow}   \\

    & RMSE$\downarrow$ & PSNR$\uparrow$ & SSIM$\uparrow$ & LPIPS$\downarrow$
    & RMSE$\downarrow$ & PSNR$\uparrow$ & SSIM$\uparrow$
    & RMSE$\downarrow$ & PSNR$\uparrow$ & SSIM$\uparrow$
    \\
    
    \midrule
    \midrule
    							
    (a) Feature addition
    & 3.545	& 33.76	& 0.9582 & 0.0548 
    & 6.021	& 38.32	& 0.9893
    & 3.060	& 36.77	& 0.9755
    \\
    							
    (b) Feature multiplication
    & 3.798	& 32.99	& 0.9554 & 0.0574 
    & 6.879	& 37.22	& 0.9882
    & 3.195	& 36.41	& 0.9744
    \\
					
    (c) Feature concatenation 
    & 3.513	& 33.91 & 0.9595 & 0.0530 
    & 6.121	& \topone{38.37} & 0.9895
    & 3.002	& 37.04 & 0.9764
    \\
								
    (d) SF w/o soft-mask
    & 3.528	& 33.62	& 0.9585 & 0.0542 
    & 6.293	& 38.09	& 0.9892
    & 2.987	& 36.69	& 0.9756
    \\
    \midrule
														
    (e) w/o SAB-$\mathbf{1}$  
    & 3.543	& 33.66	& 0.9580 & 0.0540 
    & 6.246	& 38.27	& 0.9894
    & 3.013	& 36.70	& 0.9750
    \\
														
    (f) w/o SAB-$\mathbf{2}$   
    & 3.599	& 33.77	& 0.9589 & 0.0536 
    & 6.549	& 38.26	& 0.9891
    & 3.022	& 37.00	& 0.9764
    \\
														
    (g) w/o SAB-$\mathbf{L-1}$   
    & 3.689	& 33.76	& 0.9570 & 0.0571
    & 6.314	& 38.26	& 0.9889
    & 3.175	& 36.82	& 0.9744
    \\
    \midrule
							
    (h) One encoder-decoder  
    & 3.911	& 33.06	& 0.9545 & 0.0638 
    & 7.319	& 37.04	& 0.9875
    & 3.243	& 36.59	& 0.9743
    \\
    							
    (i) Two encoder-decoder  
    & 3.855	& 33.01	& 0.9554 & 0.0619 
    & 7.227	& 36.96	& 0.9880
    & 3.194	& 36.54	& 0.9748
    \\
    \midrule
							
    (j) Iteration-1
    & 3.451	& 33.57	& 0.9598 & 0.0541 
    & 6.555	& 37.52	& 0.9890
    & \topone{2.844}	& \topone{37.23}	& \topone{0.9778}
    \\

    (k) Iteration-2
    & 3.474	& 33.60	& 0.9589 & 0.0541 
    & 6.308	& 37.91	& 0.9894
    & 2.919	& 36.87	& 0.9762
    \\
    							
    (l) Iteration-3
    & 3.460	& 33.78	& 0.9591 & 0.0538 
    & 6.236	& 37.99	& 0.9894
    & 2.916	& 37.03	& 0.9765
    \\
    							
    \topone{(m) Iteration-4(Ours)} 
    & \topone{3.401} & \topone{33.94} & \topone{0.9598} & \topone{0.0529} 
    & \topone{5.938} & 38.28 & \topone{0.9896}
    & 2.904 & 37.04 & 0.9765
    \\

    \bottomrule

	\end{tabular}
	}
	}
}
\label{tab:ablation_istd+}
\vspace{-10pt}
\end{table*}

In this section, we conduct comprehensive ablation experiments to validate each part of the proposed method, and the results are displayed in \tableref{tab:ablation_istd+}. 

\subsubsection{Effectiveness of SAB}
Firstly, we evaluate the effectiveness of the smart aggregation block by comparing it with three substituted aggregation operations: feature addition, which adds $\mathbf{F}_l$ and $\hat{\mathbf{F}}_l$; feature multiplication, which multiplies $\mathbf{F}_l$ and $\hat{\mathbf{F}}_l$; and feature concatenation, which concatenates $\mathbf{F}_l$ and $\hat{\mathbf{F}}_l$. 
Empowered by the smart aggregation block, our method achieves the highest reconstruction quality across all evaluation metrics, as demonstrated in (a)-(c). 
Specifically, at the whole image level, replacing the smart aggregation block with feature addition leads to an increase of 4.23\% in RMSE and 3.59\% in LPIPS, and a decrease of 0.53\% in PSNR and 0.17\% in SSIM. 
Replacing the smart aggregation block with feature multiplication leads to an increase of 11.67\% in RMSE and 8.51\% in LPIPS, and a decrease of 2.80\% in PSNR and 0.46\% in SSIM. 
Replacing the smart aggregation block with feature concatenation leads to an increase of 3.29\% in RMSE and 0.19\% in LPIPS, and a decrease of 0.09\% in PSNR and 0.03\% in SSIM. 

To further demonstrate the necessity of each smart aggregation block, we conduct experiments to remove them individually. As shown in (e)-(g), we find that removing any of the smart aggregation blocks leads to a decrease in reconstruction quality across all metrics. 
At the whole image level, removing the SAB after the first layer leads to an increase of 4.18\% in RMSE and 2.08\% in LPIPS, and a decrease of 0.82\% in PSNR and 0.19\% in SSIM. 
Removing the SAB after the third layer leads to an increase of 5.82\% in RMSE and 1.32\% in LPIPS, and a decrease of 0.50\% in PSNR and 0.09\% in SSIM. 
Removing the SAB after the second-to-last layer leads to an increase of 8.47\% in RMSE and 7.94\% in LPIPS, and a decrease of 0.53\% in PSNR and 0.29\% in SSIM.

\subsubsection{Effectiveness of the soft mask}
Besides, we evaluate the significance of the soft mask $\hat{\mathbf{M}}_l$ produced in the smart aggregation block by removing it directly. As shown in (d), we observe that without $\hat{\mathbf{M}}_l$, the reconstruction quality significantly deteriorates. Specifically, it leads to an increase of 3.73\% in RMSE and 2.46\% in LPIPS, and a decrease of 0.94\% in PSNR and 0.14\% in SSIM at the whole image level. 

\subsubsection{Effectiveness of the Dual-Branch shadow removal paradigm}
Furthermore, we compare our method with an encoder-decoder architecture in two scenarios. Firstly, we evaluate a single saved model that is optimal for the whole image. As shown in (h), our method outperformed this scenario. Specifically, at the whole image level, our method achieves a reduction of 13.04\% in RMSE and 17.08\% in LPIPS, as well as an increase of 2.66\% in PSNR and 0.56\% in SSIM. 
Secondly, we evaluated two saved models that are optimal for the shadow regions and non-shadow regions, respectively. To obtain a de-shadowed clean image, we can combine the restored results of these two selected models using the binary shadow masks.
As shown in (i), our method also outperforms this scenario. Specifically, at the whole image level, our method achieves a reduction of 11.78\% in RMSE and 14.54\% in LPIPS, as well as an increase of 2.82\% in PSNR and 0.46\% in SSIM. 

\subsubsection{Effectiveness of Iterative}
Finally, we evaluate the performance of our method by iterating at different times. As shown in (j)-(m), increasing the number of iterations significantly improves performance in the shadow regions, with only a negligible decrease in performance in the non-shadow regions. 
Specifically, compared to iteration-1, our method reduces the RMSE in shadow regions by 9.41\% while only increasing it in non-shadow regions by 2.11\%.
Compared to iteration-2, our method surprisingly reduces the RMSE in both shadow and non-shadow regions by 5.87\% and 0.51\%, respectively. 
Similarly, in comparison to iteration-3, our method achieves a reduction in RMSE of 4.78\% and 0.41\% in shadow and non-shadow regions, respectively.

\section{Conclusions}
In this work, we first identify the limitation of existing shadow removal approaches that use a shared model to restore both shadow and non-shadow regions. 
To overcome this limitation, we propose to decouple shadow removal into two distinct tasks: restoring shadow regions to their shadow-free counterparts and identical mapping for non-shadow regions.
Specifically, our proposed method comprises three components. Firstly, we employ an identical mapping branch (IMB) to handle the non-shadow regions. 
Secondly, we use an iterative de-shadow branch (IDB) to handle the shadow regions by progressively transferring information from the non-shadow regions to the shadow regions in an iterative manner, which facilitates the process of shadow removal. 
Finally, we design a smart aggregation block (SAB) to adaptive integrate features from both IMB and IDB.
The extensive experiments demonstrate the superiority of our proposed method over all state-of-the-art competitors.
\bibliographystyle{ACM-Reference-Format}
\bibliography{ref}


\begin{thebibliography}{45}


\ifx \showCODEN    \undefined \def \showCODEN     #1{\unskip}     \fi
\ifx \showDOI      \undefined \def \showDOI       #1{#1}\fi
\ifx \showISBNx    \undefined \def \showISBNx     #1{\unskip}     \fi
\ifx \showISBNxiii \undefined \def \showISBNxiii  #1{\unskip}     \fi
\ifx \showISSN     \undefined \def \showISSN      #1{\unskip}     \fi
\ifx \showLCCN     \undefined \def \showLCCN      #1{\unskip}     \fi
\ifx \shownote     \undefined \def \shownote      #1{#1}          \fi
\ifx \showarticletitle \undefined \def \showarticletitle #1{#1}   \fi
\ifx \showURL      \undefined \def \showURL       {\relax}        \fi
\providecommand\bibfield[2]{#2}
\providecommand\bibinfo[2]{#2}
\providecommand\natexlab[1]{#1}
\providecommand\showeprint[2][]{arXiv:#2}

\bibitem[Abiko and Ikehara(2022)]%
        {abiko2022channel}
\bibfield{author}{\bibinfo{person}{Ryo Abiko} {and} \bibinfo{person}{Masaaki
  Ikehara}.} \bibinfo{year}{2022}\natexlab{}.
\newblock \showarticletitle{Channel attention GAN trained with enhanced dataset
  for single-image shadow removal}.
\newblock \bibinfo{journal}{\emph{IEEE Access}}  \bibinfo{volume}{10}
  (\bibinfo{year}{2022}), \bibinfo{pages}{12322--12333}.
\newblock


\bibitem[Cun et~al\mbox{.}(2020)]%
        {cun2020towards}
\bibfield{author}{\bibinfo{person}{Xiaodong Cun}, \bibinfo{person}{Chi-Man
  Pun}, {and} \bibinfo{person}{Cheng Shi}.} \bibinfo{year}{2020}\natexlab{}.
\newblock \showarticletitle{Towards ghost-free shadow removal via dual
  hierarchical aggregation network and shadow matting GAN}. In
  \bibinfo{booktitle}{\emph{Proceedings of the AAAI Conference on Artificial
  Intelligence}}, Vol.~\bibinfo{volume}{34}. \bibinfo{pages}{10680--10687}.
\newblock


\bibitem[Finlayson et~al\mbox{.}(2009)]%
        {finlayson2009entropy}
\bibfield{author}{\bibinfo{person}{Graham~D Finlayson}, \bibinfo{person}{Mark~S
  Drew}, {and} \bibinfo{person}{Cheng Lu}.} \bibinfo{year}{2009}\natexlab{}.
\newblock \showarticletitle{Entropy minimization for shadow removal}.
\newblock \bibinfo{journal}{\emph{International Journal of Computer Vision}}
  \bibinfo{volume}{85}, \bibinfo{number}{1} (\bibinfo{year}{2009}),
  \bibinfo{pages}{35--57}.
\newblock


\bibitem[Finlayson et~al\mbox{.}(2005)]%
        {finlayson2005removal}
\bibfield{author}{\bibinfo{person}{Graham~D Finlayson},
  \bibinfo{person}{Steven~D Hordley}, \bibinfo{person}{Cheng Lu}, {and}
  \bibinfo{person}{Mark~S Drew}.} \bibinfo{year}{2005}\natexlab{}.
\newblock \showarticletitle{On the removal of shadows from images}.
\newblock \bibinfo{journal}{\emph{IEEE transactions on pattern analysis and
  machine intelligence}} \bibinfo{volume}{28}, \bibinfo{number}{1}
  (\bibinfo{year}{2005}), \bibinfo{pages}{59--68}.
\newblock


\bibitem[Fu et~al\mbox{.}(2021a)]%
        {fu2021benchmarking}
\bibfield{author}{\bibinfo{person}{Lan Fu}, \bibinfo{person}{Qing Guo},
  \bibinfo{person}{Felix Juefei-Xu}, \bibinfo{person}{Hongkai Yu},
  \bibinfo{person}{Wei Feng}, \bibinfo{person}{Yang Liu}, {and}
  \bibinfo{person}{Song Wang}.} \bibinfo{year}{2021}\natexlab{a}.
\newblock \showarticletitle{Benchmarking shadow removal for facial landmark
  detection and beyond}.
\newblock \bibinfo{journal}{\emph{arXiv preprint arXiv:2111.13790}}
  (\bibinfo{year}{2021}).
\newblock


\bibitem[Fu et~al\mbox{.}(2021b)]%
        {fu2021deep}
\bibfield{author}{\bibinfo{person}{Lan Fu}, \bibinfo{person}{Hongkai Yu},
  \bibinfo{person}{Xiaoguang Li}, \bibinfo{person}{Craig~P Przybyla}, {and}
  \bibinfo{person}{Song Wang}.} \bibinfo{year}{2021}\natexlab{b}.
\newblock \showarticletitle{Deep Learning for Object Detection in
  Materials-Science Images: A tutorial}.
\newblock \bibinfo{journal}{\emph{IEEE Signal Processing Magazine}}
  \bibinfo{volume}{39}, \bibinfo{number}{1} (\bibinfo{year}{2021}),
  \bibinfo{pages}{78--88}.
\newblock


\bibitem[Fu et~al\mbox{.}(2021c)]%
        {fu2021auto}
\bibfield{author}{\bibinfo{person}{Lan Fu}, \bibinfo{person}{Changqing Zhou},
  \bibinfo{person}{Qing Guo}, \bibinfo{person}{Felix Juefei-Xu},
  \bibinfo{person}{Hongkai Yu}, \bibinfo{person}{Wei Feng},
  \bibinfo{person}{Yang Liu}, {and} \bibinfo{person}{Song Wang}.}
  \bibinfo{year}{2021}\natexlab{c}.
\newblock \showarticletitle{Auto-exposure fusion for single-image shadow
  removal}. In \bibinfo{booktitle}{\emph{Proceedings of the IEEE/CVF
  International Conference on Computer Vision}}. \bibinfo{pages}{10571--10580}.
\newblock


\bibitem[Gao et~al\mbox{.}(2022)]%
        {gao2022towards}
\bibfield{author}{\bibinfo{person}{Jianhao Gao}, \bibinfo{person}{Quanlong
  Zheng}, {and} \bibinfo{person}{Yandong Guo}.}
  \bibinfo{year}{2022}\natexlab{}.
\newblock \showarticletitle{Towards real-world shadow removal with a shadow
  simulation method and a two-stage framework}. In
  \bibinfo{booktitle}{\emph{Proceedings of the IEEE/CVF Conference on Computer
  Vision and Pattern Recognition}}. \bibinfo{pages}{599--608}.
\newblock


\bibitem[Guo et~al\mbox{.}(2021)]%
        {guo2021jpgnet}
\bibfield{author}{\bibinfo{person}{Qing Guo}, \bibinfo{person}{Xiaoguang Li},
  \bibinfo{person}{Felix Juefei-Xu}, \bibinfo{person}{Hongkai Yu},
  \bibinfo{person}{Yang Liu}, {and} \bibinfo{person}{Song Wang}.}
  \bibinfo{year}{2021}\natexlab{}.
\newblock \showarticletitle{JPGNet: Joint Predictive Filtering and Generative
  Network for Image Inpainting}. In \bibinfo{booktitle}{\emph{Proceedings of
  the 29th ACM International Conference on Multimedia}}.
  \bibinfo{pages}{386--394}.
\newblock


\bibitem[Guo et~al\mbox{.}(2012)]%
        {guo2012paired}
\bibfield{author}{\bibinfo{person}{Ruiqi Guo}, \bibinfo{person}{Qieyun Dai},
  {and} \bibinfo{person}{Derek Hoiem}.} \bibinfo{year}{2012}\natexlab{}.
\newblock \showarticletitle{Paired regions for shadow detection and removal}.
\newblock \bibinfo{journal}{\emph{IEEE transactions on pattern analysis and
  machine intelligence}} \bibinfo{volume}{35}, \bibinfo{number}{12}
  (\bibinfo{year}{2012}), \bibinfo{pages}{2956--2967}.
\newblock


\bibitem[Hang et~al\mbox{.}(2019)]%
        {hang2019cascaded}
\bibfield{author}{\bibinfo{person}{Renlong Hang}, \bibinfo{person}{Qingshan
  Liu}, \bibinfo{person}{Danfeng Hong}, {and} \bibinfo{person}{Pedram
  Ghamisi}.} \bibinfo{year}{2019}\natexlab{}.
\newblock \showarticletitle{Cascaded recurrent neural networks for
  hyperspectral image classification}.
\newblock \bibinfo{journal}{\emph{IEEE Transactions on Geoscience and Remote
  Sensing}} \bibinfo{volume}{57}, \bibinfo{number}{8} (\bibinfo{year}{2019}),
  \bibinfo{pages}{5384--5394}.
\newblock


\bibitem[He et~al\mbox{.}(2021)]%
        {he2021unsupervised}
\bibfield{author}{\bibinfo{person}{Yingqing He}, \bibinfo{person}{Yazhou Xing},
  \bibinfo{person}{Tianjia Zhang}, {and} \bibinfo{person}{Qifeng Chen}.}
  \bibinfo{year}{2021}\natexlab{}.
\newblock \showarticletitle{Unsupervised Portrait Shadow Removal via Generative
  Priors}. In \bibinfo{booktitle}{\emph{Proceedings of the 29th ACM
  International Conference on Multimedia}}. \bibinfo{pages}{236--244}.
\newblock


\bibitem[Hu et~al\mbox{.}(2019a)]%
        {hu2019direction}
\bibfield{author}{\bibinfo{person}{Xiaowei Hu}, \bibinfo{person}{Chi-Wing Fu},
  \bibinfo{person}{Lei Zhu}, \bibinfo{person}{Jing Qin}, {and}
  \bibinfo{person}{Pheng-Ann Heng}.} \bibinfo{year}{2019}\natexlab{a}.
\newblock \showarticletitle{Direction-aware spatial context features for shadow
  detection and removal}.
\newblock \bibinfo{journal}{\emph{IEEE TPAMI}} \bibinfo{volume}{42},
  \bibinfo{number}{11} (\bibinfo{year}{2019}), \bibinfo{pages}{2795--2808}.
\newblock


\bibitem[Hu et~al\mbox{.}(2019b)]%
        {hu2019maskshadowgan}
\bibfield{author}{\bibinfo{person}{Xiaowei Hu}, \bibinfo{person}{Yitong Jiang},
  \bibinfo{person}{Chi-Wing Fu}, {and} \bibinfo{person}{Pheng-Ann Heng}.}
  \bibinfo{year}{2019}\natexlab{b}.
\newblock \showarticletitle{Mask-ShadowGAN: Learning to remove shadows from
  unpaired data}. In \bibinfo{booktitle}{\emph{Proceedings of the IEEE/CVF
  International Conference on Computer Vision}}. \bibinfo{pages}{2472--2481}.
\newblock


\bibitem[Inoue and Yamasaki(2020)]%
        {inoue2020learning}
\bibfield{author}{\bibinfo{person}{Naoto Inoue} {and}
  \bibinfo{person}{Toshihiko Yamasaki}.} \bibinfo{year}{2020}\natexlab{}.
\newblock \showarticletitle{Learning from synthetic shadows for shadow
  detection and removal}.
\newblock \bibinfo{journal}{\emph{IEEE Transactions on Circuits and Systems for
  Video Technology}} \bibinfo{volume}{31}, \bibinfo{number}{11}
  (\bibinfo{year}{2020}), \bibinfo{pages}{4187--4197}.
\newblock


\bibitem[Jin et~al\mbox{.}(2021)]%
        {jin2021dc}
\bibfield{author}{\bibinfo{person}{Yeying Jin}, \bibinfo{person}{Aashish
  Sharma}, {and} \bibinfo{person}{Robby~T Tan}.}
  \bibinfo{year}{2021}\natexlab{}.
\newblock \showarticletitle{DC-ShadowNet: Single-Image Hard and Soft Shadow
  Removal Using Unsupervised Domain-Classifier Guided Network}. In
  \bibinfo{booktitle}{\emph{Proceedings of the IEEE/CVF International
  Conference on Computer Vision}}. \bibinfo{pages}{5027--5036}.
\newblock


\bibitem[Johnson et~al\mbox{.}(2016)]%
        {johnson2016perceptual}
\bibfield{author}{\bibinfo{person}{Justin Johnson}, \bibinfo{person}{Alexandre
  Alahi}, {and} \bibinfo{person}{Li Fei-Fei}.} \bibinfo{year}{2016}\natexlab{}.
\newblock \showarticletitle{Perceptual losses for real-time style transfer and
  super-resolution}. In \bibinfo{booktitle}{\emph{European conference on
  computer vision}}. Springer, \bibinfo{pages}{694--711}.
\newblock


\bibitem[Le and Samaras(2019)]%
        {le2019shadow}
\bibfield{author}{\bibinfo{person}{Hieu Le} {and} \bibinfo{person}{Dimitris
  Samaras}.} \bibinfo{year}{2019}\natexlab{}.
\newblock \showarticletitle{Shadow removal via shadow image decomposition}. In
  \bibinfo{booktitle}{\emph{Proceedings of the IEEE/CVF International
  Conference on Computer Vision}}. \bibinfo{pages}{8578--8587}.
\newblock


\bibitem[Le and Samaras(2020)]%
        {le2020from}
\bibfield{author}{\bibinfo{person}{Hieu Le} {and} \bibinfo{person}{Dimitris
  Samaras}.} \bibinfo{year}{2020}\natexlab{}.
\newblock \showarticletitle{From shadow segmentation to shadow removal}. In
  \bibinfo{booktitle}{\emph{European Conference on Computer Vision}}. Springer,
  \bibinfo{pages}{264--281}.
\newblock


\bibitem[Li et~al\mbox{.}(2020)]%
        {li2020recurrent}
\bibfield{author}{\bibinfo{person}{Jingyuan Li}, \bibinfo{person}{Ning Wang},
  \bibinfo{person}{Lefei Zhang}, \bibinfo{person}{Bo Du}, {and}
  \bibinfo{person}{Dacheng Tao}.} \bibinfo{year}{2020}\natexlab{}.
\newblock \showarticletitle{Recurrent feature reasoning for image inpainting}.
  In \bibinfo{booktitle}{\emph{Proceedings of the IEEE/CVF Conference on
  Computer Vision and Pattern Recognition}}. \bibinfo{pages}{7760--7768}.
\newblock


\bibitem[Li et~al\mbox{.}(2022b)]%
        {li2022practical}
\bibfield{author}{\bibinfo{person}{Jiankun Li}, \bibinfo{person}{Peisen Wang},
  \bibinfo{person}{Pengfei Xiong}, \bibinfo{person}{Tao Cai},
  \bibinfo{person}{Ziwei Yan}, \bibinfo{person}{Lei Yang},
  \bibinfo{person}{Jiangyu Liu}, \bibinfo{person}{Haoqiang Fan}, {and}
  \bibinfo{person}{Shuaicheng Liu}.} \bibinfo{year}{2022}\natexlab{b}.
\newblock \showarticletitle{Practical stereo matching via cascaded recurrent
  network with adaptive correlation}. In \bibinfo{booktitle}{\emph{Proceedings
  of the IEEE/CVF Conference on Computer Vision and Pattern Recognition}}.
  \bibinfo{pages}{16263--16272}.
\newblock


\bibitem[Li et~al\mbox{.}(2023)]%
        {li2023leveraging}
\bibfield{author}{\bibinfo{person}{Xiaoguang Li}, \bibinfo{person}{Qing Guo},
  \bibinfo{person}{Rabab Abdelfattah}, \bibinfo{person}{Di Lin},
  \bibinfo{person}{Wei Feng}, \bibinfo{person}{Ivor Tsang}, {and}
  \bibinfo{person}{Song Wang}.} \bibinfo{year}{2023}\natexlab{}.
\newblock \showarticletitle{Leveraging Inpainting for Single-Image Shadow
  Removal}.
\newblock \bibinfo{journal}{\emph{arXiv preprint arXiv:2302.05361}}
  (\bibinfo{year}{2023}).
\newblock


\bibitem[Li et~al\mbox{.}(2022a)]%
        {li2022misf}
\bibfield{author}{\bibinfo{person}{Xiaoguang Li}, \bibinfo{person}{Qing Guo},
  \bibinfo{person}{Di Lin}, \bibinfo{person}{Ping Li}, \bibinfo{person}{Wei
  Feng}, {and} \bibinfo{person}{Song Wang}.} \bibinfo{year}{2022}\natexlab{a}.
\newblock \showarticletitle{MISF: Multi-level Interactive Siamese Filtering for
  High-Fidelity Image Inpainting}. In \bibinfo{booktitle}{\emph{Proceedings of
  the IEEE/CVF Conference on Computer Vision and Pattern Recognition}}.
  \bibinfo{pages}{1869--1878}.
\newblock


\bibitem[Liu et~al\mbox{.}(2021a)]%
        {liu2021shadow}
\bibfield{author}{\bibinfo{person}{Zhihao Liu}, \bibinfo{person}{Hui Yin},
  \bibinfo{person}{Yang Mi}, \bibinfo{person}{Mengyang Pu}, {and}
  \bibinfo{person}{Song Wang}.} \bibinfo{year}{2021}\natexlab{a}.
\newblock \showarticletitle{Shadow removal by a lightness-guided network with
  training on unpaired data}.
\newblock \bibinfo{journal}{\emph{IEEE Transactions on Image Processing}}
  \bibinfo{volume}{30} (\bibinfo{year}{2021}), \bibinfo{pages}{1853--1865}.
\newblock


\bibitem[Liu et~al\mbox{.}(2021b)]%
        {liu2021from}
\bibfield{author}{\bibinfo{person}{Zhihao Liu}, \bibinfo{person}{Hui Yin},
  \bibinfo{person}{Xinyi Wu}, \bibinfo{person}{Zhenyao Wu},
  \bibinfo{person}{Yang Mi}, {and} \bibinfo{person}{Song Wang}.}
  \bibinfo{year}{2021}\natexlab{b}.
\newblock \showarticletitle{From Shadow Generation to Shadow Removal}. In
  \bibinfo{booktitle}{\emph{Proceedings of the IEEE/CVF Conference on Computer
  Vision and Pattern Recognition}}. \bibinfo{pages}{4927--4936}.
\newblock


\bibitem[Mohan et~al\mbox{.}(2007)]%
        {mohan2007editing}
\bibfield{author}{\bibinfo{person}{Ankit Mohan}, \bibinfo{person}{Jack
  Tumblin}, {and} \bibinfo{person}{Prasun Choudhury}.}
  \bibinfo{year}{2007}\natexlab{}.
\newblock \showarticletitle{Editing soft shadows in a digital photograph}.
\newblock \bibinfo{journal}{\emph{IEEE Computer Graphics and Applications}}
  \bibinfo{volume}{27}, \bibinfo{number}{2} (\bibinfo{year}{2007}),
  \bibinfo{pages}{23--31}.
\newblock


\bibitem[Nadimi and Bhanu(2004)]%
        {nadimi2004physical}
\bibfield{author}{\bibinfo{person}{Sohail Nadimi} {and} \bibinfo{person}{Bir
  Bhanu}.} \bibinfo{year}{2004}\natexlab{}.
\newblock \showarticletitle{Physical models for moving shadow and object
  detection in video}.
\newblock \bibinfo{journal}{\emph{IEEE transactions on pattern analysis and
  machine intelligence}} \bibinfo{volume}{26}, \bibinfo{number}{8}
  (\bibinfo{year}{2004}), \bibinfo{pages}{1079--1087}.
\newblock


\bibitem[Qu et~al\mbox{.}(2017)]%
        {qu2017deshadownet}
\bibfield{author}{\bibinfo{person}{Liangqiong Qu}, \bibinfo{person}{Jiandong
  Tian}, \bibinfo{person}{Shengfeng He}, \bibinfo{person}{Yandong Tang}, {and}
  \bibinfo{person}{Rynson~WH Lau}.} \bibinfo{year}{2017}\natexlab{}.
\newblock \showarticletitle{Deshadownet: A multi-context embedding deep network
  for shadow removal}. In \bibinfo{booktitle}{\emph{Proceedings of the IEEE
  Conference on Computer Vision and Pattern Recognition}}.
  \bibinfo{pages}{4067--4075}.
\newblock


\bibitem[Saharia et~al\mbox{.}(2022)]%
        {saharia2022image}
\bibfield{author}{\bibinfo{person}{Chitwan Saharia}, \bibinfo{person}{Jonathan
  Ho}, \bibinfo{person}{William Chan}, \bibinfo{person}{Tim Salimans},
  \bibinfo{person}{David~J Fleet}, {and} \bibinfo{person}{Mohammad Norouzi}.}
  \bibinfo{year}{2022}\natexlab{}.
\newblock \showarticletitle{Image super-resolution via iterative refinement}.
\newblock \bibinfo{journal}{\emph{IEEE Transactions on Pattern Analysis and
  Machine Intelligence}} (\bibinfo{year}{2022}).
\newblock


\bibitem[Sanin et~al\mbox{.}(2010)]%
        {sanin2010improved}
\bibfield{author}{\bibinfo{person}{Andres Sanin}, \bibinfo{person}{Conrad
  Sanderson}, {and} \bibinfo{person}{Brian~C Lovell}.}
  \bibinfo{year}{2010}\natexlab{}.
\newblock \showarticletitle{Improved shadow removal for robust person tracking
  in surveillance scenarios}. In \bibinfo{booktitle}{\emph{ICPR}}.
  \bibinfo{pages}{141--144}.
\newblock


\bibitem[Wan et~al\mbox{.}(2022)]%
        {wan2022sg-shadow}
\bibfield{author}{\bibinfo{person}{Jin Wan}, \bibinfo{person}{Hui Yin},
  \bibinfo{person}{Zhenyao Wu}, \bibinfo{person}{Xinyi Wu},
  \bibinfo{person}{Yanting Liu}, {and} \bibinfo{person}{Song Wang}.}
  \bibinfo{year}{2022}\natexlab{}.
\newblock \showarticletitle{Style-Guided Shadow Removal}. In
  \bibinfo{booktitle}{\emph{Proceedings of the European Conference on Computer
  Vision (ECCV)}}.
\newblock


\bibitem[Wang et~al\mbox{.}(2022)]%
        {wang2022yolov7}
\bibfield{author}{\bibinfo{person}{Chien-Yao Wang}, \bibinfo{person}{Alexey
  Bochkovskiy}, {and} \bibinfo{person}{Hong-Yuan~Mark Liao}.}
  \bibinfo{year}{2022}\natexlab{}.
\newblock \showarticletitle{YOLOv7: Trainable bag-of-freebies sets new
  state-of-the-art for real-time object detectors}.
\newblock \bibinfo{journal}{\emph{arXiv preprint arXiv:2207.02696}}
  (\bibinfo{year}{2022}).
\newblock


\bibitem[Wang and Wang(2022)]%
        {wang2022iterative}
\bibfield{author}{\bibinfo{person}{Dong Wang} {and} \bibinfo{person}{Xiao-Ping
  Wang}.} \bibinfo{year}{2022}\natexlab{}.
\newblock \showarticletitle{The iterative convolution--thresholding method
  (ICTM) for image segmentation}.
\newblock \bibinfo{journal}{\emph{Pattern Recognition}}  \bibinfo{volume}{130}
  (\bibinfo{year}{2022}), \bibinfo{pages}{108794}.
\newblock


\bibitem[Wang et~al\mbox{.}(2021)]%
        {wang2021pac}
\bibfield{author}{\bibinfo{person}{Zifeng Wang}, \bibinfo{person}{Shao-Lun
  Huang}, \bibinfo{person}{Ercan~E Kuruoglu}, \bibinfo{person}{Jimeng Sun},
  \bibinfo{person}{Xi Chen}, {and} \bibinfo{person}{Yefeng Zheng}.}
  \bibinfo{year}{2021}\natexlab{}.
\newblock \showarticletitle{PAC-bayes information bottleneck}.
\newblock \bibinfo{journal}{\emph{arXiv preprint arXiv:2109.14509}}
  (\bibinfo{year}{2021}).
\newblock


\bibitem[Wei et~al\mbox{.}(2019)]%
        {wei2019shadow}
\bibfield{author}{\bibinfo{person}{Jinjiang Wei}, \bibinfo{person}{Chengjiang
  Long}, \bibinfo{person}{Hua Zou}, {and} \bibinfo{person}{Chunxia Xiao}.}
  \bibinfo{year}{2019}\natexlab{}.
\newblock \showarticletitle{Shadow inpainting and removal using generative
  adversarial networks with slice convolutions}. In
  \bibinfo{booktitle}{\emph{Computer Graphics Forum}},
  Vol.~\bibinfo{volume}{38}. Wiley Online Library, \bibinfo{pages}{381--392}.
\newblock


\bibitem[Yan et~al\mbox{.}(2022)]%
        {yan2022rignet}
\bibfield{author}{\bibinfo{person}{Zhiqiang Yan}, \bibinfo{person}{Kun Wang},
  \bibinfo{person}{Xiang Li}, \bibinfo{person}{Zhenyu Zhang},
  \bibinfo{person}{Jun Li}, {and} \bibinfo{person}{Jian Yang}.}
  \bibinfo{year}{2022}\natexlab{}.
\newblock \showarticletitle{RigNet: Repetitive image guided network for depth
  completion}. In \bibinfo{booktitle}{\emph{Computer Vision--ECCV 2022: 17th
  European Conference, Tel Aviv, Israel, October 23--27, 2022, Proceedings,
  Part XXVII}}. Springer, \bibinfo{pages}{214--230}.
\newblock


\bibitem[Yu et~al\mbox{.}(2019)]%
        {yu2019deep}
\bibfield{author}{\bibinfo{person}{Songhyun Yu}, \bibinfo{person}{Bumjun Park},
  {and} \bibinfo{person}{Jechang Jeong}.} \bibinfo{year}{2019}\natexlab{}.
\newblock \showarticletitle{Deep iterative down-up cnn for image denoising}. In
  \bibinfo{booktitle}{\emph{Proceedings of the IEEE/CVF conference on computer
  vision and pattern recognition workshops}}. \bibinfo{pages}{0--0}.
\newblock


\bibitem[Zhan and Lu(2019)]%
        {zhan2019esir}
\bibfield{author}{\bibinfo{person}{Fangneng Zhan} {and}
  \bibinfo{person}{Shijian Lu}.} \bibinfo{year}{2019}\natexlab{}.
\newblock \showarticletitle{Esir: End-to-end scene text recognition via
  iterative image rectification}. In \bibinfo{booktitle}{\emph{Proceedings of
  the IEEE/CVF conference on computer vision and pattern recognition}}.
  \bibinfo{pages}{2059--2068}.
\newblock


\bibitem[Zhang et~al\mbox{.}(2020)]%
        {zhang2020ris}
\bibfield{author}{\bibinfo{person}{Ling Zhang}, \bibinfo{person}{Chengjiang
  Long}, \bibinfo{person}{Xiaolong Zhang}, {and} \bibinfo{person}{Chunxia
  Xiao}.} \bibinfo{year}{2020}\natexlab{}.
\newblock \showarticletitle{Ris-gan: Explore residual and illumination with
  generative adversarial networks for shadow removal}. In
  \bibinfo{booktitle}{\emph{Proceedings of the AAAI Conference on Artificial
  Intelligence}}, Vol.~\bibinfo{volume}{34}. \bibinfo{pages}{12829--12836}.
\newblock


\bibitem[Zhang et~al\mbox{.}(2015)]%
        {zhang2015shadow}
\bibfield{author}{\bibinfo{person}{Ling Zhang}, \bibinfo{person}{Qing Zhang},
  {and} \bibinfo{person}{Chunxia Xiao}.} \bibinfo{year}{2015}\natexlab{}.
\newblock \showarticletitle{Shadow remover: Image shadow removal based on
  illumination recovering optimization}.
\newblock \bibinfo{journal}{\emph{IEEE Transactions on Image Processing}}
  \bibinfo{volume}{24}, \bibinfo{number}{11} (\bibinfo{year}{2015}),
  \bibinfo{pages}{4623--4636}.
\newblock


\bibitem[Zhang et~al\mbox{.}(2018a)]%
        {zhang2018unreasonable}
\bibfield{author}{\bibinfo{person}{Richard Zhang}, \bibinfo{person}{Phillip
  Isola}, \bibinfo{person}{Alexei~A Efros}, \bibinfo{person}{Eli Shechtman},
  {and} \bibinfo{person}{Oliver Wang}.} \bibinfo{year}{2018}\natexlab{a}.
\newblock \showarticletitle{The unreasonable effectiveness of deep features as
  a perceptual metric}. In \bibinfo{booktitle}{\emph{IEEE/CVF Conference on
  Computer Vision and Pattern Recognition (CVPR)}}. \bibinfo{pages}{586--595}.
\newblock


\bibitem[Zhang et~al\mbox{.}(2018b)]%
        {zhang2018improving}
\bibfield{author}{\bibinfo{person}{Wuming Zhang}, \bibinfo{person}{Xi Zhao},
  \bibinfo{person}{Jean-Marie Morvan}, {and} \bibinfo{person}{Liming Chen}.}
  \bibinfo{year}{2018}\natexlab{b}.
\newblock \showarticletitle{Improving shadow suppression for illumination
  robust face recognition}.
\newblock \bibinfo{journal}{\emph{IEEE transactions on pattern analysis and
  machine intelligence}} \bibinfo{volume}{41}, \bibinfo{number}{3}
  (\bibinfo{year}{2018}), \bibinfo{pages}{611--624}.
\newblock


\bibitem[Zhou et~al\mbox{.}(2022)]%
        {zhou2022edge}
\bibfield{author}{\bibinfo{person}{Xiaofei Zhou}, \bibinfo{person}{Kunye Shen},
  \bibinfo{person}{Li Weng}, \bibinfo{person}{Runmin Cong},
  \bibinfo{person}{Bolun Zheng}, \bibinfo{person}{Jiyong Zhang}, {and}
  \bibinfo{person}{Chenggang Yan}.} \bibinfo{year}{2022}\natexlab{}.
\newblock \showarticletitle{Edge-guided recurrent positioning network for
  salient object detection in optical remote sensing images}.
\newblock \bibinfo{journal}{\emph{IEEE Transactions on Cybernetics}}
  \bibinfo{volume}{53}, \bibinfo{number}{1} (\bibinfo{year}{2022}),
  \bibinfo{pages}{539--552}.
\newblock


\bibitem[Zhu et~al\mbox{.}(2022a)]%
        {zhu2022bijective}
\bibfield{author}{\bibinfo{person}{Yurui Zhu}, \bibinfo{person}{Jie Huang},
  \bibinfo{person}{Xueyang Fu}, \bibinfo{person}{Feng Zhao},
  \bibinfo{person}{Qibin Sun}, {and} \bibinfo{person}{Zheng-Jun Zha}.}
  \bibinfo{year}{2022}\natexlab{a}.
\newblock \showarticletitle{Bijective Mapping Network for Shadow Removal}. In
  \bibinfo{booktitle}{\emph{Proceedings of the IEEE/CVF Conference on Computer
  Vision and Pattern Recognition}}. \bibinfo{pages}{5627--5636}.
\newblock


\bibitem[Zhu et~al\mbox{.}(2022b)]%
        {zhu2022efficient}
\bibfield{author}{\bibinfo{person}{Yurui Zhu}, \bibinfo{person}{Zeyu Xiao},
  \bibinfo{person}{Yanchi Fang}, \bibinfo{person}{Xueyang Fu},
  \bibinfo{person}{Zhiwei Xiong}, {and} \bibinfo{person}{Zheng-Jun Zha}.}
  \bibinfo{year}{2022}\natexlab{b}.
\newblock \showarticletitle{Efficient Model-Driven Network for Shadow Removal}.
\newblock  (\bibinfo{year}{2022}).
\newblock


\end{thebibliography}

\end{document}